\documentclass{article}

\usepackage{microtype}
\usepackage{graphicx}
\usepackage{subcaption}
\usepackage{booktabs} 

\usepackage{graphicx}
\usepackage{subcaption} 
\usepackage[table]{xcolor}

\usepackage{hyperref}

\usepackage[preprint]{icml2026}

\usepackage{amsmath}
\usepackage{amssymb}
\usepackage{bbm} 
\usepackage{mathtools}
\usepackage{amsthm}
\usepackage{xcolor}    
\usepackage{etoolbox}
\usepackage{multirow}
\usepackage{makecell}

\usepackage{algorithm}
\usepackage{algorithmic}

\usepackage[capitalize,noabbrev]{cleveref}

\theoremstyle{plain}
\newtheorem{theorem}{Theorem}[section]

\newtheorem{lemma}[theorem]{Lemma}

\theoremstyle{definition}

\newtheorem{assumption}[theorem]{Assumption}

\theoremstyle{example}
\newtheorem{example}[theorem]{Example}
\usepackage[most]{tcolorbox} 

\tcbset{
  formula box/.style={
    enhanced,
    left=4pt,
  }
}

\usepackage[textsize=tiny]{todonotes}

\icmltitlerunning{Submission and Formatting Instructions for ICML 2026}

\begin{document}

\twocolumn[
  \icmltitle{Large-Scale Auto-bidding with Nash Equilibrium Constraints}



  \icmlsetsymbol{equal}{*}

  \begin{icmlauthorlist}
    \icmlauthor{Zhiyu Mou}{yyy}
    \icmlauthor{Miao Xu}{yyy}
    \icmlauthor{Ronquan Bai}{yyy}
    \icmlauthor{Zhuoran Yang}{sch}
    \icmlauthor{Chuan Yu}{yyy}
    \icmlauthor{Jian Xu}{yyy}
    \icmlauthor{Bo Zheng}{yyy}
  \end{icmlauthorlist}

  \icmlaffiliation{yyy}{Alibaba Group, Beijing,
China.}
  \icmlaffiliation{sch}{Department of Statistics
and Data Science, Yale University, New Haven, US. }


  \icmlkeywords{auto-bidding, deep learning, nash equilibrium, online advertising}

  \vskip 0.3in
]



 \printAffiliationsAndNotice{}  

\begin{abstract}
Auto-bidding has become a cornerstone of modern online advertising platforms, enabling many advertisers to automate bidding at scale and optimize campaign performance.
However, prevailing industrial systems rely on single-agent auto-bidding methods that are scalable but overlook the strategic interdependence among advertisers’ bids, leading to unstable or suboptimal outcomes.
While recent works recognize the game-theoretic nature of auto-bidding, existing approaches remain either computationally intractable at scale or lack a principled equilibrium-selection that aligns with platform-wide objectives.
In this paper, we bridge this gap by introducing \emph{Nash Equilibrium-Constrained Bidding} (NCB), a principled and scalable auto-bidding framework that recasts auto-bidding as a platform-wide optimization problem subject to Nash equilibrium constraints.
This approach accounts for fine-grained strategic interdependencies among advertisers, ensuring both agent-level stability and ecosystem-level optimality.
Notably, we develop a theoretically sound penalty-based primal-dual gradient method with rigorous convergence guarantees, supported by an efficient algorithm suitable for industrial deployment. 
Extensive experiments validate the effectiveness of our approach. 
\end{abstract}

\section{Introduction}
\label{sec:intro}
In online advertising, auto-bidding has become one of the most widely adopted approaches for advertisers to compete for impression opportunities. It allows advertisers to specify only high-level objectives during bidding, eliminating the need to adjust bids manually and thereby substantially reducing their operational burden \cite{adikari2015real}.
Many online advertising platforms, such as Google and Alibaba, offer auto-bidding services, in which each advertiser is represented by a dedicated, self-interested agent that bids on their behalf to optimize advertising performance \cite{deng2024efficiency}.
The effectiveness of these agents hinges critically on the underlying auto-bidding algorithms, which jointly governs individual advertisers' outcomes and platform-level metrics, such as Gross Merchandise Volume (GMV).

\begin{figure}[t]
    \centering
    \includegraphics[width=1.0\linewidth]{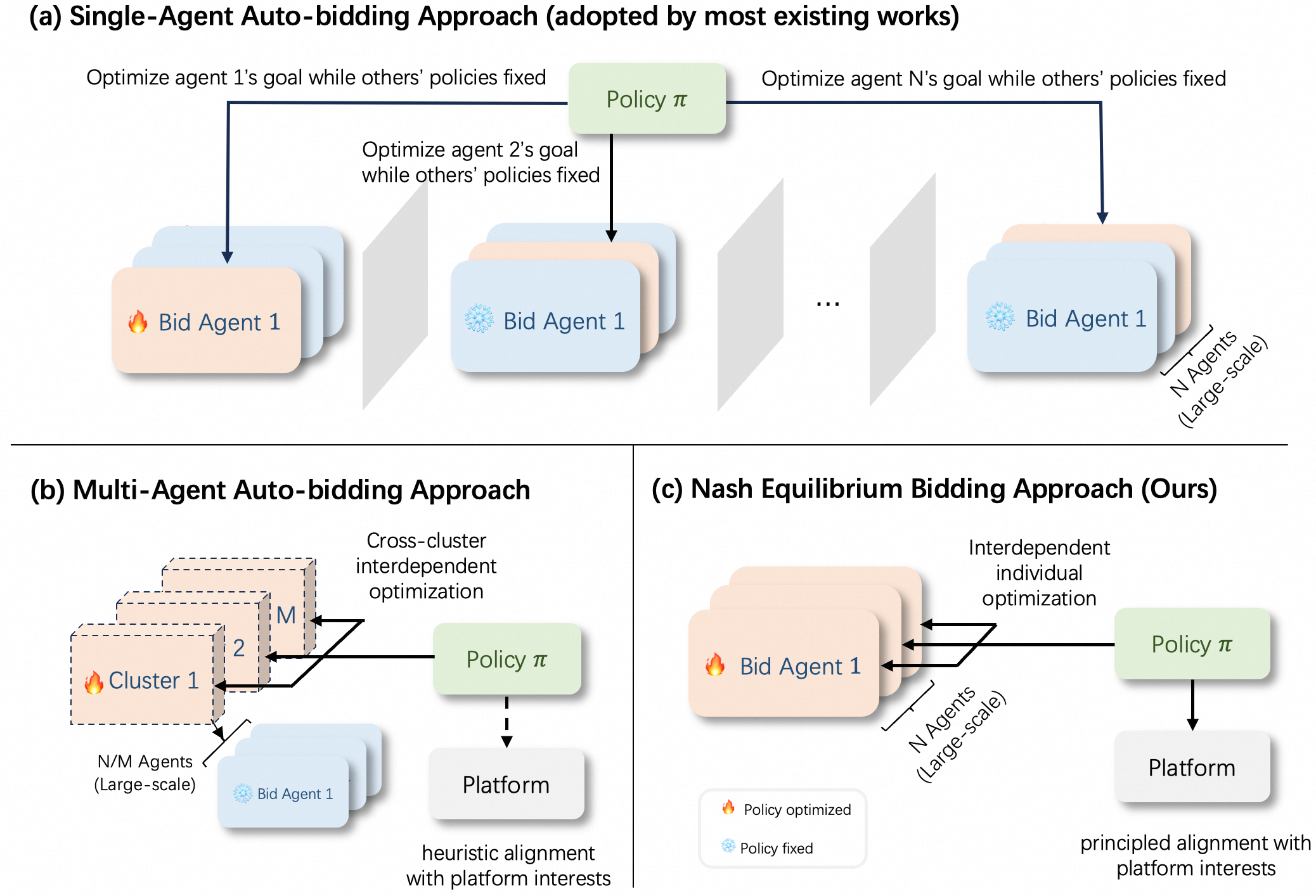}
    \caption{Illustrations of conventional auto-bidding approaches versus the proposed framework for large-scale industrial systems. The proposed Nash-Equilibrium Bidding (NCB) approach accounts for both strategic agent interactions and the principled interplay between advertisers' incentives and the platform's interests.}
    \label{fig:auto_bidding_process}
\end{figure}

Recent advancements in auto-bidding algorithms for industrial advertising systems predominantly adopt a single-agent perspective \cite{USCB, SORL, SAB_1, guo2024generative, SAB_3,mou2025enhancing}, optimizing each agent's bidding policy while treating other agents' policies as fixed, as shown in Fig.\ref{fig:auto_bidding_process}(a).
This modeling choice is primarily driven by the need for scalability in real-world advertising markets, where the large number of advertisers (thousands or even millions) renders vanilla joint optimization over all agents computationally intractable. However, it entirely neglects mutual influence among agents, resulting in weaker equilibrium guarantees and suboptimal advertiser outcomes. Moreover, these methods also fail to capture collective effects on platform-level metrics, limiting system-wide optimality.

While some studies investigate multi-agent approaches for industrial auto-bidding systems \cite{MAAB, DCMAB}, they typically cluster agents into a few coarse groups and model only inter-cluster interactions, failing to capture intra-cluster strategic dependencies in a rigorous or game-theoretically consistent manner, as shown in Fig.\ref{fig:auto_bidding_process}(b). This often results in suboptimal bidding policies and distorted equilibrium outcomes within groups. Moreover, these approaches lack a principled problem formulation that explicitly and coherently captures the coupling between advertiser-level utilities and platform-wide metrics. Consequently, they often resort to heuristic methods without provable guarantees for equilibrium convergence or system-wide performance.
In summary, a principled and scalable auto-bidding framework that jointly accounts for fine-grained strategic interdependencies among agents and the platform’s ecosystem-level goals remains an open challenge.



{In parallel, recent theoretical research in auto-bidding has reframed the problem as a multi-agent strategic interaction among rational, self-interested agents.
This line of work investigates pacing equilibria \cite{conitzer2022multiplicative, conitzer2022pacing}, a state where no advertiser can improve their utility by unilaterally deviating from their pacing strategy. 
While foundational studies establish the existence, and potential multiplicitym of such equilibria, their computational methods are typically restricted to small-scale scenarios with dozens of agents, precluding their application in real-world markets comprising thousands or even millions of advertisers.
Moreover, another bottleneck of these works lies in the absence of principled equilibrium selection mechanisms.
They only ensure convergence to some Nash equilibrium, without any guarantee of reaching a preferred equilibrium (e.g., one that maximizes social welfare or platform revenue) among the potentially many that exist.
See Appendix \ref{app:related_work} for detailed related work discussion.
}

{
In summary, practical methods prioritize scalability by ignoring strategic interactions among campaigns, while theoretical equilibrium analyses are typically either non-scalable or offer no control over equilibrium selection, limiting their direct applicability in large-scale industrial systems.
}

To bridging this gap, we propose \emph{Nash Equilibrium Constrained Bidding} (NCB), a principled and scalable auto-bidding framework that accounts for fine-grained strategic interdependencies among agents while jointly optimizing platform-wide metrics.
Specifically, as shown in Fig. \ref{fig:auto_bidding_process}(c), we reframe the auto-bidding problem through a novel prospective that simultaneously considers agent-level strategic equilibria and platform-wide objectives, grounded in two core principles: (1) since each agent is self-interested and aims to optimize their advertiser’s performance, any stable outcome corresponds to an Nash equilibrium (NE); (2) among the set of such equilibria\footnote{Note that multiple NE exist \cite{wang2002reinforcement}.}, we select the one that best aligns with the platform's ecosystem-level objectives, such as total GMV.
Consequently, we formulate the auto-bidding problem as the maximization of platform metrics subject to the NE constraint. However, it appears a high-dimensional, bi-level optimization problem, which is hard to solve.

Notably, we propose a theoretical approach for the formulated auto-bidding problem with rigorous convergence guarantees. 
Specifically, we reformulate the original high-dimensional, bi-level optimization as a constrained problem over agents’ bidding factors, leveraging the structure of their best-response strategies. To handle the resulting non-smooth NE constraints, we apply the Fischer-Burmeister function to obtain a smooth and differentiable surrogate. We then develop a penalty-based primal-dual gradient framework with analytically tractable gradients, enabling efficient solution of the constrained optimization. Under mild, practically motivated assumptions, we provide theoretical results establishing that our algorithm converges globally to a Karush–Kuhn–Tucker (KKT) point of the original problem, thereby ensuring both equilibrium stability and social welfare optimality.

Moreover, we provide an efficient calculation pipeline of the theoretical approach and rigorously analyze its computational complexity.
Notably, by exploiting the problem’s structure and pre-computing key intermediate quantities, we reduce the per-iteration time and space complexity, making the proposed method scalable to large-scale industrial settings.

Furthermore, to account for the impression dynamic nature in real-world auto-bidding systems,  we develop a practical deep learning-based algorithm that adaptively predicts optimal bids based on sequentially observed impression features. This approach enables efficient online decision-making without requiring full knowledge of future impressions, ensuring that the system remains robust and responsive to changing market conditions.
Extensive experiments valid the effectiveness of our approach. 




\section{Preliminaries}
\label{sec:pre}
We consider $N\in\mathbb{N}_+$ advertisers competing for $K\in \mathbb{N}_+$ impressions through the platform's auto-bidding service, where both $N$ and $K$ are typically very large. 
Each advertiser $i \in [N]$ seeks to maximize the total value of their winning impressions subject to a budget constraint $B_i\in\mathbb{R}_+$.
To facilitate this, the platform assigns each advertiser $i$ an auto-bidding agent that manages bids on their behalf, optimizing for the advertiser's total winning value.
Let $b_{i,k}\in\mathbb{R}_+$ denote the bid of agent $i$ for impression $k\in[K]$, and let $v_{i,k}\in [0, V]$ represent the impression value. Here, $V\in\mathbb{R}_+$ denotes the value upper bound.

\textbf{Auction Mechanism.}
In a standard second-price auction, agent $i$ wins impression $k$ if and only if their bid $b_{i,k}$ exceeds the highest competing bid $m_{i,k}\triangleq \max_{j\neq  i} b_{j,k}$. The allocation is thus deterministic and given by the indicator $\mathbbm{1}\{b_{i,k}> m_{i,k}\}$, where ties are broken arbitrarily. The winning agent $i$ pays $m_{i,k}$.
However, real-world implementations often deviate from this idealized model due to operational constraints, such as resource-smoothing policies or frequency capping, which introduce stochasticity into the allocation process. 
To account for this uncertainty, we adopt a soft allocation rule in which the probability $p_{i,k}$ of agent $i$ winning impression $k$ is modeled as a softmax over all bids, i.e., $ p_{i,k}\triangleq\frac{\exp{(b_{i,k}/\tau)}}{ \sum_j \exp{(b_{j,k}/\tau)}}$, where $\tau$ is a small positive. 
Note that $p_{i,k}$ approaches the deterministic allocation rule $\mathbbm{1}\{b_{i,k}> m_{i,k}\}$ as $\tau\rightarrow 0$, and its sum over all agent $i$ is $1$, i.e., $\sum_i p_{i,k}=1$.
Consequently, the market price is modeled as the weighted sum of other agents' bids, i.e., $m_{i,k}\triangleq \sum_{j\neq i} p^{-i}_{j,k}b_{j,k}$, where $p^{-i}_{j,k}$ denotes the winning probability of agent $j$ among agents except agent $i$, i.e., 
\begin{align}
    p^{-i}_{j,k}\triangleq\frac{\exp(b_{j,k}/\tau)}{\sum_{i'\neq i}\exp(b_{i',k}/\tau)}=\frac{p_{j,k}}{1-p_{i,k}}.
\end{align}

The effectiveness of the agents hinges critically on the bid set $\{b_{i,k}\}_{i\in[N],k\in[K]}$, which jointly governs each advertiser's expected return and platform-level metrics. 


\textbf{Unilateral Best Response.}
Due to the difficulty of dealing with strategic interactions among a large number of agents, prevailing auto-bidding algorithms adopt a single-agent approximation \cite{SAB_1, SAB_3, SORL, USCB, guo2024generative}.
Specifically, as illustrated in Fig.~\ref{fig:auto_bidding_process}(a), these algorithms optimize each agent’s bids in isolation, assuming that the bids of all other agents remain fixed.
This approach inherently seeks each agent's unilateral best response.
Formally, it can be formulated as: $\forall i$,
\begin{align}
\label{equ:single_agent_problem}
&\max_{b_i}\;\;\underbrace{\sum_{k\in[K]}p_{i,k}(b_i;b_{-i})v_{i,k}}_{\triangleq R_i(b_i;b_{-i})},\notag\\
&\;\mathrm{s.t.} \;\;\underbrace{\sum_{k\in[K]}p_{i,k}(b_i;b_{-i})m_{i,k}(b_i;b_{-i})}_{\triangleq C_i(b_i;b_{-i})}\le B_i, 
\end{align}
where $b_{i}\triangleq \{b_{i,k}\}_{k=1}^K$ denotes the bids of agent $i$ and $b_{-i}$ denotes the bids of agents other than agent $i$,  $R_i(b_i;b_{-i})$ and $C_i(b_i;b_{-i})$ represent the total value and cost of agent $i$, respectively.
As is well established in Theorem 2.1 in \cite{USCB}, the solution to Eq.~\eqref{equ:single_agent_problem} is to bid proportionally to the value of each impression, i.e.,
\begin{align}
\label{equ:bid_stru}
    b_{i,k}=\alpha_i v_{i,k},\quad \forall i,k
\end{align}
where $\alpha_{i}\in [0, A]$ is referred to as the \emph{bidding factor} and is identical across all impressions for each agent $i$, and $A\in\mathbb{R}_+$ denotes its upper bound. 
Under the bidding structure in Eq.~\eqref{equ:bid_stru}, both the aggregate value and the total cost for each agent $i$ are proportional to $\alpha_i$.
Consequently, we can obtain the following optimality condition for $\alpha_i$, and the complete proof is provided in Appendix \ref{app:proof_lemma}.
\begin{lemma}[Unilateral Best Response Condition]
    \label{lemma:opt}
    The optimal bidding factor $\alpha_i$ for agent $i$'s best-response problem in Eq.~\eqref{equ:single_agent_problem} either exhausts agent $i$'s budget or saturates at the upper bound $A$, whichever occurs first, i.e., 
    \begin{align}
        C_i(\alpha_iv_i;b_{-i})=\min(B_i,C_i(Av_i;b_{-i})),
    \end{align}
    where $v_i\triangleq[v_{i,1},v_{i,2},\cdots,v_{i,K}]$.
\end{lemma}

However, the single-agent approach entirely overlooks how other agents’ policies evolve during optimization, resulting in deficient equilibrium guarantees and suboptimal outcomes for individual advertisers.
Moreover, it fails to account for the collective impact of all agents on platform-level metrics, limiting its system-wide optimality.

In summary, a principled and scalable auto-bidding framework that jointly accounts for fine-grained strategic interdependencies among agents and the platform’s ecosystem-level goals remains an open challenge.

\section{Nash Equilibrium Constrained Bidding}
\label{sec:method}
In this section, we introduce \emph{Nash Equilibrium Constrained Bidding} (NCB). We present a principled formulation of the auto-bidding problem that jointly accounts for fine-grained strategic interdependencies among agents while simultaneously aligning with the platform's ecosystem-level objectives. 
Then, we establish a theoretical approach to the problem and provide rigorous convergence guarantees in Section \ref{sec:theoretical_approach}. 
In Section \ref{sec:practical_algo}, we translate our theoretical results into a scalable, practical algorithm tailored for the dynamic nature of impressions within industrial environments.   

\textbf{Problem Formulation.}
As each agent optimizes its bidding strategy to maximize individual advertiser performance, the resulting strategic interactions give rise to a non-cooperative game. Within this framework, the Nash Equilibrium (NE) is the canonical solution concept, characterizing a stable configuration in which no agent can unilaterally improve its utility. 
Moreover, given the potential for multiple equilibria \cite{wang2002reinforcement, paes2024complex} (see also Example \ref{example}), the platform should select the equilibrium that best aligns with its global objectives, such as maximizing social welfare or revenue. 
In this study, we define the platform's objective as the maximization of social welfare, a choice intended to foster long-term ecosystem sustainability \footnote{Furthermore, since platform revenue corresponds to the aggregate advertiser budget at the Nash Equilibrium, additional revenue optimization becomes redundant.}. 
Synthesizing these elements, we formally cast the auto-bidding problem as a social welfare maximization task subject to the NE constraint as follows:
\begin{align}
\label{equ:ncb}
&\underbrace{\max_{\{{b}_{i}\}_{i=1}^N}\;\; \sum_{i\in [N]}R_i({b}_i;{b}_{-i})	}_{\text{maximize social welfare}}\\
	&\;\;\mathrm{s.t.}\quad b_i\in\arg\bigg(\max_{b_i'}R_i(b_i';b_{-i})  \notag\\
    &\;\;\quad\quad\underbrace{\quad\quad\quad\;\;\;\;\;\;\;\;\mathrm{s.t.}\;\; C_i(b_i';b_{-i})\le B_i \bigg), \forall i}_{\text{NE constraint, the bids of each agent are the best response}}\tag{\ref{equ:ncb}{a}}\label{equ:ncb_constraint}.
\end{align}
The formulation in Eq. \eqref{equ:ncb} constitutes a high-dimensional bi-level optimization problem with $NK$ decision variables, which is non-trivial to solve.  

\subsection{Theoretical Approach}
\label{sec:theoretical_approach}

\begin{figure}[t]
    \centering
\includegraphics[width=1.0\linewidth]{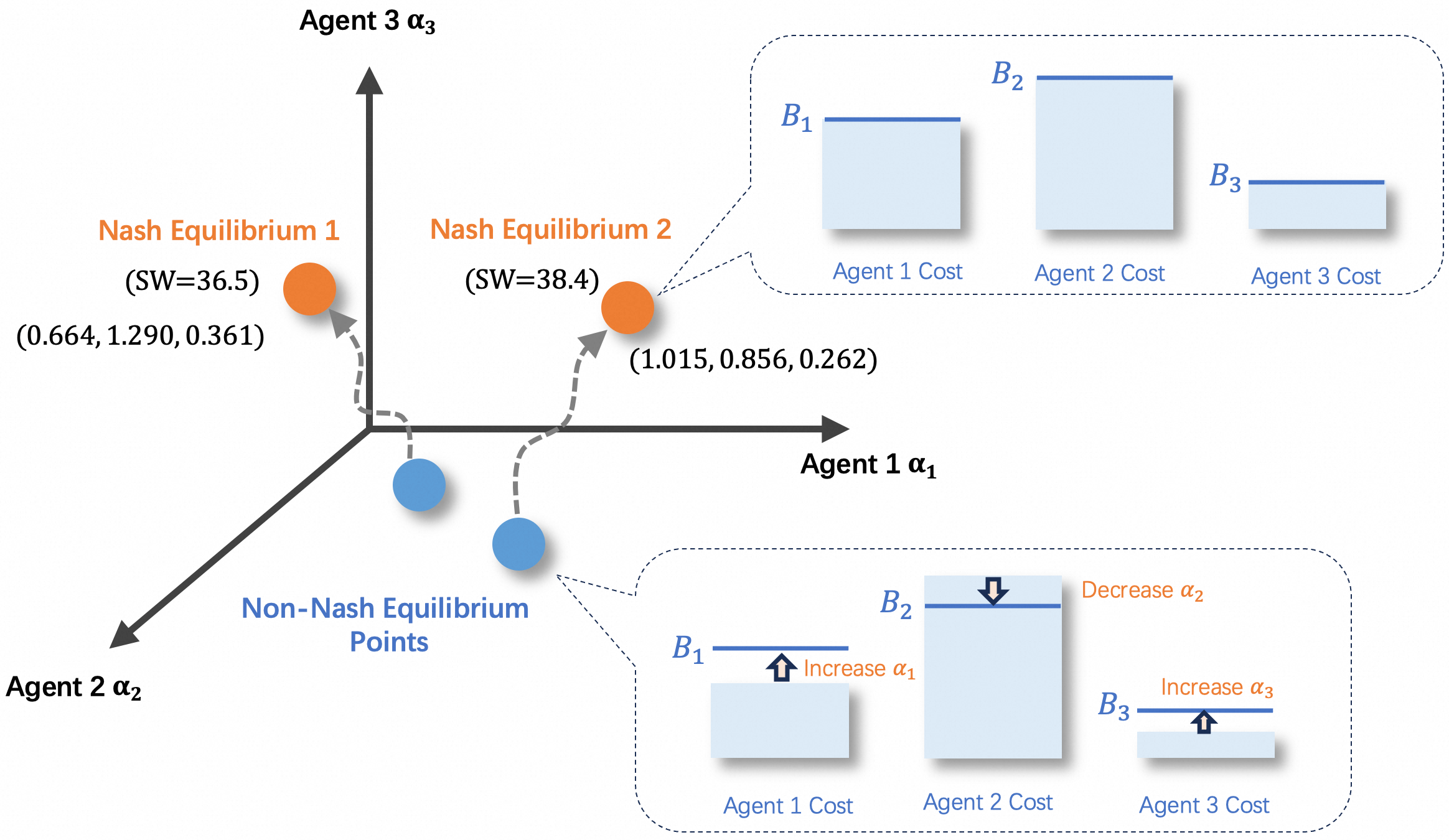}
    \caption{The strategic dynamics in illustrative Example \ref{example}. Two distinct NE points exist, both characterized by full budget exhaustion across all agents. Nevertheless, their respective social welfare outcomes exhibit a $5.2\%$ relative disparity.}
    \label{fig:ncb_game}
\end{figure}

This section presents a theoretical approach to Eq.~\eqref{equ:ncb} with rigorous convergence guarantees.  
Specifically, a key observation is that, under the NE constraint in Eq.~\eqref{equ:ncb_constraint}, the bid $b_i$ should constitute an optimal solution to agent $i$'s unilateral best response problem in Eq.~\eqref{equ:single_agent_problem}. As established previously in Eq.~\eqref{equ:bid_stru}, the optimal bidding strategy of each agent $i$ on impression $k$ takes the form of $b_{i,k}=\alpha_iv_{i,k}$, where $\alpha_i$ should satisfy Lemma \ref{lemma:opt}.
Consequently, the profile $\alpha\triangleq\{\alpha_i\}_{i=1}^N$ becomes the decision variable, and the original problem Eq.~\eqref{equ:ncb} can be equivalently reformulated as:
\begin{align}
&{\max_{\alpha\in [0,A]^N}\;\; \sum_{i\in [N]}R_i(\alpha_i;\alpha_{-i})	}\label{equ:ncb_alpha}\\
	&\;\;\mathrm{s.t.}\quad C_i(\alpha_i;\alpha_{-i})=\min(B_i, C_i(A;\alpha_{-i}))  , \forall i.\tag{\ref{equ:ncb_alpha}{a}}\label{equ:ncb_alpha_constraint}
\end{align}
However, $\min(\cdot, \cdot)$ is a non-differentiable operator, rendering the constraint set non-smooth and complicating the numerical tractability of the optimization problem.
Hence, we leverage the Fischer-Burmeister function \cite{fischer1992special} to (approximately) equivalently replace the non-differentiable constraints in Eq.~\eqref{equ:ncb_alpha_constraint}, i.e.,
\begin{align}
\label{equ:h}
    h_i(\alpha)&\triangleq B_i-C_i(\alpha_i;\alpha_{-i})+A-\alpha_i\notag\\
    &\;\;\;\;\;-\sqrt{(B_i-C_i(\alpha_i;\alpha_{-i}))^2+(A-\alpha_i)^2+\epsilon}\notag\\
    &=0,
\end{align}
where $\epsilon$ is a small positive number (e.g., $10^{-6}$) to make $h_i(\alpha)$ differentiable everywhere.
Formally, we have the following theorem that guarantees the equivalence between Eq.~\eqref{equ:h} and Eq.~\eqref{equ:ncb_alpha_constraint}, and the proof is given in Appendix \ref{app:proof_thm_equ_ne}.
\begin{theorem}[Equivalent NE Constraint]
\label{thm:equivalent_ne}
When $\epsilon=0$, $h_i(\alpha)=0,\forall i$ is equivalent to the NE constraints in Eq.~\eqref{equ:ncb_alpha}.
\end{theorem}
Therefore, the original problem in Eq.~\eqref{equ:ncb} can be equivalently reformulated as:
\begin{align}
 &{\max_{\alpha}\;\; \sum_{i\in [N]}R_i(\alpha)	}\label{equ:ncb_alpha_simplify}\\
	&\;\;\mathrm{s.t.}\quad\quad h_i(\alpha)=0,\forall i\tag{\ref{equ:ncb_alpha_simplify}{a}}\label{equ:ncb_alpha_constraint_simplify},
\end{align}
which is a constrained optimization problem with $\alpha$ as the decision variable, comprising a large-scale of $N$ variables.
Hereafterin, we use $R_i(\alpha)$ and $C_i(\alpha)$ to denote $R_i(\alpha_i;\alpha_{-i})$ and $C_i(\alpha_i;\alpha_{-i})$, respectively, for notational simplicity.

In summary, Nash Equilibrium Constrained Bidding coordinates agents to adjust their bidding factors $\alpha_i$ that exhaust their budgets or become saturated. Among all possible $\alpha$ profiles, it chooses the one with maximum social welfare. 

\textbf{Illustrative Example.}
Before deriving our solution framework, we present an illustrative example of Nash Equilibrium Constrained Bidding to elucidate its mechanics.
\begin{example}
\label{example}[See Appendix \ref{app:example} For Details]
Consider a setting with $N=3$ agents competing for $K=10$ impressions. The agents' budgets are $7.254$, $9.561$, and $0.731$, respectively. Their respective impression valuations are given in Table \ref{tab:example}.  
As shown in Table. \ref{tab:ne}, there exist two distinct NE points, where the first NE is $\alpha^*_1=(0.664, 1.290, 0.361)$ with social welfare $36.462$, and the second one is $\alpha^*_2=(1.015, 0.856, 0.262)$ with social welfare $38.368$. 
At each equilibrium, every agent exhausts its budget, while the relative difference in social welfare is \textbf{$\textbf{5.2\%}$}. 
This necessitates explicit consideration of social welfare maximization when selecting among equilibria.
Fig.~\ref{fig:ncb_game} summarizes the strategic dynamics of the example. 
\end{example}



\subsubsection{Penalty-based Gradient Framework}
To effectively and efficiently solve the problem in Eq.~\eqref{equ:ncb_alpha_simplify}, we develop a penalty-based gradient ascent framework with analytically tractable gradients.
Specifically, we construct an \emph{augmented Lagrangian function} of Eq.~\eqref{equ:ncb_alpha_simplify}, which incorporates a quadratic penalty term into the Lagrangian:
\begin{align}
    L_\rho(\alpha, \lambda)=
    \underbrace{\sum_{i\in [N]}\bigg[R_i(\alpha)+\lambda_i h_i(\alpha)\bigg]}_{\text{standard Lagrangian function}}
    -\underbrace{\frac{\rho}{2}\sum_{i\in [N]} h_i^2(\alpha)}_{\text{penalty term}},\notag
\end{align}
where $\rho>0$ is the penalty factor, $\lambda\triangleq\{\lambda_i\}_{i=1}^N\in\mathbb{R}^N$ is the Lagrangian factor.
With $L_\rho(\alpha, \lambda)$, the problem in Eq.~\eqref{equ:ncb_alpha_simplify} is subsequently recast as the max-min optimization problem, i.e., $ \max_\alpha\min_\lambda L_\rho(\alpha, \lambda)$, which is amenable to solution via primal-dual gradient iterations.
Specifically, the analytical expressions of the gradient of $L_\rho(\alpha,\lambda)$ with respect to $\alpha$ and $\lambda$ are given in Eq.~\eqref{equ:alpha_gradient} and Eq.~\eqref{equ:lambda_gradient}, respectively. Detailed derivations are given in Appendix \ref{app:proof_gradients}.

\begin{tcolorbox}[formula box]
{\small
\textbf{(1) The gradient of $L_\rho(\alpha,\lambda)$ with respect to $\alpha_j$:}
\begin{align}
\label{equ:alpha_gradient}
    \frac{\partial L_\rho(\alpha,\lambda)}{\partial\alpha_j}=\sum_{i\in [N]}\bigg[\frac{\partial R_i(\alpha)}{\partial \alpha_j}+E_i\frac{\partial h_i(\alpha)}{\partial \alpha_j}\bigg].
\end{align}
Here, $E_i\triangleq \lambda_i-\rho h_i(\alpha)$, and the return gradient is:
\begin{align}
\label{equ:R_grad}
    \frac{\partial R_i(\alpha)}{\partial \alpha_j}=\sum_{k\in [K]} v_{i,k}\frac{p_{i,k}v_{j,k}}{\tau}(\delta_{ij}-p_{j,k}),
\end{align}
where $\delta_{ij}\triangleq \mathbbm{1}\{i=j\}$ denotes the Kronecker delta.
Let $Z_i\triangleq \sqrt{(B_i-C_i(\alpha))^2+(A-\alpha_i)^2+\epsilon}$.
The expression of the constraint gradient $\frac{\partial h_i(\alpha)}{\partial \alpha_j}$ is:
\begin{align}
\label{equ:h_grad}
   \frac{\partial h_i(\alpha)}{\partial \alpha_j}= -\bigg(1-\frac{B_i-C_i}{Z_i}\bigg)\frac{\partial C_i}{\partial\alpha_j}
   -\bigg(1-\frac{A-\alpha_i}{Z_i}\bigg)\delta_{ij}\notag
\end{align}
where the cost gradient $\frac{\partial C_i(\alpha)}{\partial \alpha_j}$ is expressed as:
\begin{align}
   & \frac{\partial C_i(\alpha)}{\partial \alpha_j}=\sum_{k\in [K]}\bigg[p_{i,k}p^{-i}_{j,k}v_{j,k}\bigg[1+\frac{\alpha_j v_{j,k}-m_{i,k}}{\tau}\bigg]\notag\\
    & \quad\quad(1-\delta_{ij})+m_{i,k}\frac{p_{i,k}v_{j,k}}{\tau}(\delta_{ij}-p_{j,k})\bigg].
\end{align}
\textbf{(2) The gradient of $L_\rho(\alpha, \lambda)$ with respect to $\lambda_j$:}
\begin{align}
\label{equ:lambda_gradient}
    \frac{\partial L_\rho(\alpha,\lambda)}{\partial\lambda_j}=h_j(\alpha).
\end{align}
}
\end{tcolorbox}

Equipped with these analytical results, we solve the problem in Eq.~\eqref{equ:ncb_alpha_simplify} using primal-dual gradient iterations, which update $\alpha$ and $\lambda$ alternatively until convergence. The method details are summarized in Algorithm \ref{algo:ncb}.

\subsubsection{Theoretical Convergence Analysis}

This section establishes convergence guarantees for Algorithm~\ref{algo:ncb}. In our theoretical analysis, we assume that sufficient gradient updates are performed in the primal space such that the primal gradient approximates zero.
\begin{assumption}[Sufficient Primal Iterations]
\label{assump:primal_conv}
The primal update in Algorithm~\ref{algo:ncb} is executed until exact stationarity, i.e.,
$\nabla_\alpha L_\rho(\alpha^{t+1},\lambda^t)=0$.
\end{assumption}

Equipped with Assumption $\ref{assump:primal_conv}$, we show that if Algorithm \ref{algo:ncb} converges, then the limit point is a Karush-Kuhn-Tucker (KKT) point of the problem in Eq. \eqref{equ:ncb_alpha_simplify}.
The proof is given in Appendix \ref{app:proof_kkt_limit}.

\begin{theorem}[KKT Property of Limit Points]
\label{thm:kkt_limit}
Under Assumption \ref{assump:primal_conv}, if the sequence $\{(\alpha^t,\lambda^t)\}$ generated by Algorithm \ref{algo:ncb} converges to a certain point $(\hat{\alpha}^*,\hat{\lambda}^*)$, then this limit point admits a KKT point of the NCB problem in Eq. \eqref{equ:ncb_alpha_simplify}:
\begin{align}
 &(\text{stationary})\sum_{i\in[N]}\bigg[   \nabla_\alpha R_i(\hat{\alpha}^*)+\hat{\lambda}^*_i\nabla_\alpha h_i(\hat{\alpha}^*)\bigg]=0\notag\\
& (\text{feasibility})\;\;   h_i(\hat{\alpha}^*)=0,\forall i.\notag
\end{align}
\end{theorem}

We make a mild assumption and explain its reasonability in the following.

\begin{assumption}[Self-Dominance]
\label{assump:self_dominance}
For every agent $i$, the total cross-agent interference is no more than a fraction of the self-marginal cost, i.e., 
$\frac{\partial C_i(\alpha)}{\partial\alpha_i}>\sum_{j\neq i} |\frac{\partial C_i(\alpha)}{\partial \alpha_j}|, \forall i$. 
\end{assumption}

Note that real-world auto-bidding systems frequently exhibit sparse and localized agent interactions, particularly in large-scale markets where advertisers are heterogeneous, focusing on distinct impression segments and competing mainly against a limited set of rivals. Such structural characteristics motivate Assumption \ref{assump:self_dominance}, which formalizes the empirical tendency for an agent’s own bids to have a greater impact on its cost than those of other agents. While exceptions can arise in highly overlapping markets, this assumption captures a prevalent pattern in industrial auto-bidding environments and provides a basis for our theoretical analysis.


We present the theoretical convergence guarantee of Algorithm \ref{algo:ncb} in Theorem \ref{thm:convergence_guarantee}.
Note that $J_H(\alpha)$ is the Jacobian matrix of the constraints $[ h_1(\alpha),\cdots, h_N(\alpha)]^T$. 
It can be shown that under Assumption \ref{assump:self_dominance}, $J_H(\alpha)$ is invertible, and all its singular values are strictly positive  (See Lemma \ref{lemma:invertible_J}). 
Let $\sigma_\text{min}(J_H(\alpha))>0$ denote its smallest singular value.

\begin{theorem}[Global Convergence Guarantee]
\label{thm:convergence_guarantee}
Under Assumptions \ref{assump:primal_conv} and \ref{assump:self_dominance}, if $\sigma^2_\text{min}(J_H(\alpha))>\bar{H}_1$, then with sufficiently large penalty factor that satisfies
\begin{align}
    \rho > \frac{\Lambda_{H_0}}{\sigma^2_\text{min}(J_H(\alpha))-\bar{H}_1},
\end{align}
the sequence $\{(\alpha^t,\lambda^t)\}$ generated by Algorithm \ref{algo:ncb} is guaranteed to converge as $t \rightarrow \infty$.
Note that $\Lambda_{H_0}$ represents the maximum eigenvalue of the Hessian matrix of the Lagrangian function of Eq. \eqref{equ:ncb_alpha_simplify}, and  
$\bar{H}_1$ denotes the upper bound of $\|\sum_{i\in [N]}h_i(\alpha)\nabla^2_{\alpha\alpha}h_i(\alpha)\|_2$.
\end{theorem}
The proof is given in Appendix \ref{app:proof_conv}. 
Together with Theorem \ref{thm:kkt_limit} and  Theorem \ref{thm:convergence_guarantee}, Algorithm \ref{algo:ncb} is guaranteed to converge to the KKT points of the problem in Eq.~\eqref{equ:ncb_alpha_simplify}.

\begin{algorithm}[t]
   \caption{Nash Equilibrium Bidding Framework}
   \label{algo:ncb}
\begin{algorithmic}
{\small
   \STATE {\bfseries Input:} The penalty factor $\rho>0$, step sizes $\beta, \eta>0$, small tolerances $\epsilon_{\text{primal}}\ge0$ for bid convergence, and $\epsilon_\text{fea}\ge0$ for budget satisfaction.
   \STATE {\bfseries Output:} Solution $\hat{\alpha}^*$ to the NCB problem in Eq. \eqref{equ:ncb_alpha_simplify}.
   \STATE{\bfseries Initializations:} Initialize the values of $\alpha$ and $\lambda$, $\alpha^0$ and $\lambda^0$, iteration step $t\leftarrow 0$.
   \REPEAT
   \item[] \texttt{\textcolor{blue}{// Primal Update For $\alpha$}}
   \STATE 1: Update all $\alpha_j^t$'s with their  gradients in Eq. \eqref{equ:alpha_gradient}:
   \begin{align}
       \alpha_j^t\leftarrow\alpha_j^t + \beta \frac{\partial}{\partial \alpha_j} L_\rho(\alpha^t, \lambda^t)
   \end{align}
   until $\|\nabla_\alpha L_\rho(\alpha^t,\lambda^t)\|\le \epsilon_\text{primal}$. Let the result be $\alpha^{t+1}$.
   \item[] \texttt{\textcolor{blue}{// Dual Update For $\lambda$}}
   \STATE 2: Update $\lambda$ for one step based on the budget violation in Eq. \eqref{equ:lambda_gradient} at the new alpha $\alpha^{t+1}$:
   \begin{align}
\lambda_i^{t+1}\leftarrow\lambda_i^t - \eta\frac{\partial }{\partial\lambda_i}L_\rho(\alpha^{t+1},\lambda^t)
   \end{align}
   \STATE 3: Let 
   $t\leftarrow t+1$.
   \UNTIL{convergence, i.e., $\max_{i}|C_i(\alpha^{t})-B_i|\le \epsilon_\text{fea}$}
   \STATE Let $\hat{\alpha}^*\leftarrow\alpha^{t}$ be the solution.
   }
\end{algorithmic}
\end{algorithm}

\subsubsection{Efficient Computation Pipeline and Computational Complexity Analysis}

This section introduces a computationally efficient framework for calculating gradients in Algorithm \ref{algo:ncb}, followed by a formal analysis of its time and space complexity.

\textbf{Efficient Computation Pipeline For Algorithm \ref{algo:ncb}.}
To efficiently execute the primal update, we first pre-compute the winning probabilities $p_{i,k}$ for all agents across the impression sequence. 
Notice that for a given impression $k$, $p_{i,k}$ shares the same denominator: 
\begin{align}
\text{\colorbox{cyan!20}{$S_k$}}\triangleq\sum_{j\in[N]}\exp{(\alpha_jv_{j,k}/\tau)}
\end{align}
whose evaluation incurs a time complexity of $\mathcal{O}(N)$. Computing the denominators for the entire sequence of $K$ impressions incurs a total complexity of $\mathcal{O}(NK)$.
Each $p_{i,k}$ is subsequently calculated with an additional $O(1)$ overhead. 
Therefore, the total complexity of computing all $p_{i,k}$ is $\mathcal{O}(NK)+\mathcal{O}(NK)\times\mathcal{O}(1)=\mathcal{O}(NK)$.

We next evaluate the market prices. 
Specifically, let 
\begin{align}
    \text{\colorbox{cyan!20}{$W_k$}}\triangleq \sum_{j\in [N]}\exp(\alpha_jv_{j,k}/\tau)\alpha_jv_{j,k},
\end{align}
we compute $W_k$ for all impressions with a total complexity of $\mathcal{O}(NK)$.
Equipped with $S_{k}$ and $W_k$, the market price can be evaluated as: 
\begin{align}
    m_{i,k}=\frac{W_k-\exp(\frac{\alpha_iv_{i,k}}{\tau})\alpha_iv_{i,k}}{S_k-\exp(\frac{\alpha_i v_{i,k}}{\tau})}
\end{align}
which incurs a computational complexity of $\mathcal{O}(1)$.
Therefore, all the market prices can be calculated as with a total complexity of $\mathcal{O}(NK)+\mathcal{O}(NK)\times\mathcal{O}(1)=\mathcal{O}(NK)$.
Equipped with $\{p_{i,k}\}$ and $\{m_{i,k}\}$, all the costs $C_i$ can be evaluated with a total time complexity of $\mathcal{O}(NK)$.
\begin{tcolorbox}
    Stage 1: By pre-computing $\{S_k\}$ and $\{W_k\}$ terms, we can evaluate all winning probabilities $\{p_{i,k}\}$, market prices $\{m_{i,k}\}$, and costs $\{C_i\}$ with a total time complexity of $\mathcal{O}(NK)$.
\end{tcolorbox}

Then, we evaluate the gradients in the primal update. Specifically, evaluating the return gradients yields:
    \begin{align}
        \sum_{i\in[N]}\frac{\partial R_i(\alpha)}{\partial\alpha_j}=\sum_{k\in[K]}\bigg[\frac{p_{i,k}v_{j,k}^2}{\tau}-v_{j,k}p_{j,k}\underbrace{\sum_{i\in[N]}v_{i,k}p_{i,k}}_{\triangleq\text{\colorbox{cyan!20}{$\bar{v}_k$}}}\bigg]\notag
    \end{align}
We pre-compute {$\bar{v}_k$} for all impressions, which incurs a complexity of $\mathcal{O}(NK)$. 
Equipped with $\bar{v}_k$, the return gradient for $\alpha_j$ requires a time complexity of $\mathcal{O}(K)$, and the total time complexity of the return gradient for $\alpha$ is $\mathcal{O}(NK)$. 

Moreover, the constraint gradient $\sum_{i\in[N]}E_i\frac{\partial h_i(\alpha)}{\partial \alpha_j}$ can be further expressed as follows. The detailed derivation process is given in Appendix \ref{app:complex_derivation}.
\begin{align}
    &\sum_{i\in [N]}E_i\frac{\partial h_i(\alpha)}{\partial \alpha_j}=\sum_{k\in [K]}p_{j,k}v_{j,k}\bigg(1+\frac{\alpha_jv_{j,k}}{\tau}\bigg)I_{jk}+\notag\\
   &\sum_{k\in [K]}\bigg[a_jE_jm_{j,k}\frac{p_{j,k}v_{j,k}}{\tau}-p_{j,k}v_{j,k}\text{\colorbox{cyan!20}{$U_k$}}\bigg]+b_jE_j,
\end{align}
where $a_i \triangleq -(1-\frac{B_1-C_i}{Z_i})$, $b_i\triangleq -(1-\frac{A-\alpha_i}{Z_i})$, and
\begin{equation}
\begin{aligned}
I_{jk}\triangleq \underbrace{\sum_{i\neq j}a_iE_ip_{i,k}\frac{1}{1-p_{i,k}}}_{=\text{\colorbox{cyan!20}{$Q_k$}}-a_jE_j\frac{p_{j,k}}{1-p_{j,k}}}-\frac{1}{\tau}\underbrace{\sum_{i\neq j}a_iE_ip_{i,k}\frac{1}{1-p_{i,k}}m_{i,k}}_{=\text{\colorbox{cyan!20}{$Q_k'$}}-a_jE_j\frac{p_{j,k}}{1-p_{j,k}}m_{j,k}}\notag
\end{aligned}
\end{equation}
Here, $U_k$, $Q_k$, and $Q_k'$ are defined as:
\begin{align}
     & \text{\colorbox{cyan!20}{$U_k$}}\triangleq \sum_{i\in [N]}a_iE_im_{i,k}\frac{p_{i,k}}{\tau},\\
     &\text{\colorbox{cyan!20}{$Q_k$}}\triangleq\sum_{i\in [N]} a_iE_i\frac{p_{i,k}}{1-p_{i,k}}, \\ &\text{\colorbox{cyan!20}{$Q'_k$}}\triangleq\sum_{i\in [N]} a_iE_i\frac{p_{i,k}}{1-p_{i,k}}m_{i,k}
\end{align}
Notably, by pre-computing $\{U_k\}$, $\{Q_k\}$, and $\{Q_k'\}$ for all impressions, the computational complexity of the constraint gradient  $\sum_{i\in[N]}E_i\frac{\partial h_i(\alpha)}{\partial \alpha_j}$ is reduced to $\mathcal{O}(K)$. Therefore, the total computational complexity is $\mathcal{O}(NK)$.
\begin{tcolorbox}
    Stage 2: By pre-computing $\{\bar{v}_k\}$, $\{U_k\}$, $\{Q_k\}$, and $\{Q_k'\}$ terms, we can evaluate the return gradients and constraint gradients for all agents with a total time complexity of $\mathcal{O}(NK)$.
\end{tcolorbox}
Denote the number of steps for primal update convergence as $T_p\in\mathbb{N}_+$.
The total time complexity of the primal update is $\mathcal{O}(T_pNK)$.
For the dual update, $h_j(\alpha)$ can be directly calculated with the pre-calculated $C_j$ using a time complexity of $\mathcal{O}(1)$. Therefore, the dual update computation complexity is $\mathcal{O}(N)$.
\begin{tcolorbox}
Stage 3: Iteratively execute Stages 1 and 2 until convergence. The primal update requires a total of $\mathcal{O}(T_pNK)$ time. The dual update is performed with $\mathcal{O}(N)$ time complexity.
\end{tcolorbox}
\textbf{Overall Time Complexity.}
Let $T_{A}\in \mathbb{N}_{+}$ denote the total number of primal-dual iterations required for convergence. 
The total time complexity of Algorithm \ref{algo:ncb} is $\mathcal{O}(T_AT_pNK)$.

\textbf{Overall Space Complexity.}
During the execution of Algorithm \ref{algo:ncb}, storing $\{S_k,W_k,\bar{v}_k,U_k,Q_k,Q_k'\}$ incurs a space complexity of $\mathcal{O}(K)$, while the storage of $\alpha$, $\lambda$ and $\{C_i\}$ entails $\mathcal{O}(N)$ each.
Moreover, the storage of $\{p_{i,k}, m_{i,k}\}$ requires $\mathcal{O}(NK)$ each. 
In summary, the total space complexity of Algorithm \ref{algo:ncb} is $\mathcal{O}(NK)$

\subsection{Practical Algorithm}
\label{sec:practical_algo}

This section focuses on the practical design of algorithms for large-scale auto-bidding systems. 
A significant challenge in practical implementations is that impressions are revealed sequentially, whereas Algorithm 1 requires complete knowledge of impressions in advance. 
While one could periodically invoke Algorithm 1 using historical data from recent bidding episodes \cite{aggarwal2019autobidding}, the highly dynamic nature of real-world impressions necessitates frequent online re-execution to adapt to shifting patterns. 
Because Algorithm 1's computational complexity scales as O(NK), frequent re-execution becomes computationally prohibitive for real-time online systems.

\begin{figure}
    \centering
    \includegraphics[width=1.0\linewidth]{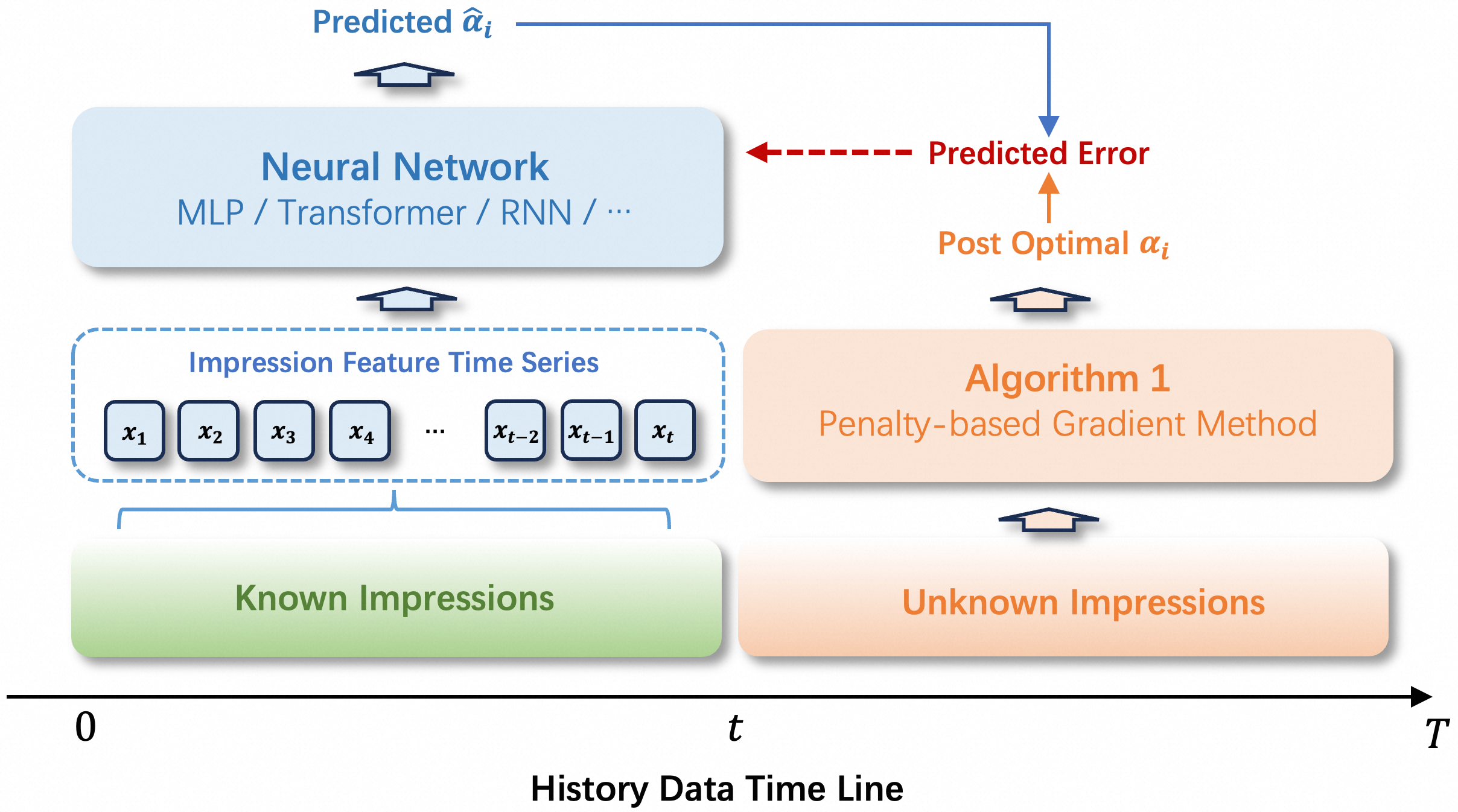}
    \caption{Practical algorithm design for large-scale auto-bidding. }
    \label{fig:practical_algo}
    \vspace{-0.1cm}
\end{figure}

To address these critical challenges, we propose a practical deep learning-based algorithm that is both computationally efficient for online deployment and does not require any prior knowledge of incoming impressions. 
Our basic idea is to train a neural network $\mathcal{M}$ to predict the optimal $\alpha$ profile for incoming impressions based on the characteristics of previously observed impressions. 
Specifically, as illustrated in Fig.~\ref{fig:practical_algo}, upon the arrival of $K_1<K$ impressions, the neural network ingests the features of these $K_1$ arrivals to predict the optimal $\alpha$ profile for the remaining $K-K_1$ incoming impressions.
During training, the target $\alpha$ profile is derived via Algorithm \ref{algo:ncb} using each agent's remaining budget and full knowledge of all incoming impressions, yielding a hindsight-optimal supervisory signal.
Notably, this design introduces an information asymmetry between the neural network's inputs and its predictive target, encouraging the model to internalize the underlying distribution of the entire sequence.
In this way, we effectively integrate both the future impression prediction and the subsequent optimal $\alpha$ calculation into a single, end-to-end differentiable inference step.

\textbf{Model Architecture.}
Note that our method is architecture-agnostic, accommodating various neural network backbones for encoding the sequence of observed impressions, including MLPs, Transformers, and RNNs.
The model's inputs are the features of impressions received for agent $i$, denoted $x_i$. 
The features can include a broad range of arrival characteristics, such as impression count, click rate, conversion rate, etc. 
The model predicts the optimal $\alpha_i$ of agent $i$ for the remaining impressions, i.e., $\alpha_i=\mathcal{M}(x_i)$.
The predicted $\alpha$ profile can be obtained by performing inference $N$ times, one for each agent, which yielding $\alpha=\{\mathcal{M}(x_i)\}_{i=1}^N$.

\begin{table*}[t]
	\caption{Performances of algorithms under different $\rho$ settings (mean $\pm$ std) based on an open source advertising system \cite{su2024auctionnet}.
    Note that the Max Exploitability is normalized by Social Welfare.
    }
	\label{table:simulated_exp}
	
	\begin{center}
		\begin{small}
				\begin{tabular}{cccccc}
					\toprule
					\textbf{Algorithms}&\textbf{Social Welfare}&\textbf{Max Exploitability}  & \textbf{$\rho$ Setting}& \textbf{Compliance Rate} & \textbf{Revenue} \\
					\toprule
					\renewcommand\arraystretch{0.8}
					SAB \cite{USCB} &4722.5$\pm$42.2 &0.166$\pm$0.0021 &/&/&5130.2$\pm$152.3\\
					\midrule
						\renewcommand\arraystretch{0.8}
					MAAB \cite{MAAB}&4880.1$\pm$96.4&0.157$\pm$0.0022&/&/&4854.5$\pm$236.5\\
					\midrule
						\renewcommand\arraystretch{0.8}
					DCMAB \cite{DCMAB}&4761.6$\pm$104.7&0.138$\pm$0.0041&/&/&5009.4$\pm$436.1\\
					\midrule
                    \renewcommand\arraystretch{0.8}
					ES \cite{jothimurugan2022specification}&4836.2$\pm$85.1&0.145$\pm$0.0015&/&/&5110.4$\pm$225.9\\
					\midrule
                    \rowcolor{cyan!10}
		&\textbf{5572.3}$\pm$63.8&	\textbf{0.140}$\pm$0.0042&	\textbf{10}&	\textbf{66.18\%}	&4670.4$\pm$264.1\\
     \rowcolor{cyan!10}
	\textbf{NCB}	&\textbf{5302.5}$\pm$35.8	&\textbf{0.134}$\pm$0.0038	&\textbf{20}&	\textbf{71.35\%}	&4942.7$\pm$145.1\\
     \rowcolor{cyan!10}
	 \textbf{Framework}	&\textbf{4979.9}$\pm$51.8&	\textbf{0.098}$\pm$0.0049&	\textbf{50}&	\textbf{76.71\%}	&5291.3$\pm$183.5\\
     \rowcolor{cyan!10}
	 \textbf{(ours)}	&	{4832.3}$\pm$46.7	&\textbf{0.063}$\pm$0.0143&	\textbf{80}&	\textbf{78.25\%}	&5327.3$\pm$269.1\\
     \rowcolor{cyan!10}
		&	4692.5$\pm$63.5	&\textbf{0.052}$\pm$0.0079	&\textbf{100}	&\textbf{80.14\%}	&5478.5$\pm$239.2
		\\
					\bottomrule
				\end{tabular}
			\end{small}
			
		\end{center}
	\vspace{-5mm}
	\end{table*}

\textbf{Training Stage.}
To construct the training dataset, we aggregate a comprehensive corpus of $D\in\mathbb{N}_+$ historical bidding trajectories $\tau_d$ from the past, denoted as $\mathcal{D}\triangleq\{\tau_d\}_{d=1}^{D}$. Each trajectory $\tau_d$ comprises the full sequence of $K_d\in\mathbb{N}_+$ impressions, including their respective values for each participating agent.
During training, we uniformly sample a trajectory $\tau_d$ from $\mathcal{D}$ and stochastically select a point of time where $K_d'<K_d$ impressions have arrived. The model is then optimized via the following objective function:
\begin{align}
    l_\mathcal{M}=\mathbb{E}_{\tau_d\sim \mathcal{D}}\bigg[\frac{1}{N}\sum_{i\in [N]}\bigg[\alpha_{i,d}-\mathcal{M}(x_{i,d})\bigg]^2\bigg],
\end{align}
where $x_{i,d}$ denotes the features of the $K_d'$ observed impressions for agent $i$. 
The ground-truth targets $\{\alpha_{i,d}\}_{i=1}^N$ are computed via Algorithm \ref{algo:ncb}, utilizing full hindsight knowledge of the remaining $K_d-K_d'$ impressions and the residual budgets of all agents.
The model is optimized using Adam until the objective function $l_\mathcal{M}$ converges.

\textbf{Online Inference.}
To account for the dynamic nature of the impressions during online deployment, we partition the whole bidding episode into $U\in\mathbb{N}_+$ discrete time steps. At each step, the model adaptively recalibrates the bidding factor profile $\alpha$ based on historical observations, ensuring the bidding strategy remains responsive to real-time fluctuations.
The computational complexity at each time step is $\mathcal{O}(NI)$, where $I\in\mathbb{N}_+$ denotes the neural network's inference overhead.
Since $I$ is independent of the number of impressions, our approach ensures that the recalibration process remains highly scalable for real-time bidding applications as impression volume increases.

\section{Experiments}
\label{sec:experiments}
We conduct experiments to validate the effectiveness of our approach. 
Specifically, the experiments are conducted based on an open source advertising system \cite{su2024auctionnet}
with 1000 agents and 70,000 impressions. Note that each algorithm is tested with 10 random seeds to obtain the standard deviation (std) of the performance.

\textbf{Performance Metric.} 
In our experiments, we focus on \emph{Social Welfare} as the primary objective, using \emph{Max Exploitability} and \emph{Compliance Rate} to assess the satisfaction of the NE constraint.
Specifically, the Max Exploitability is defined as 
\begin{align}
\max_{i\in[N]} \frac{\max_{\alpha}R_i(\alpha)-R_i(\alpha)}{\sum_{i\in[N]}R_i(\alpha)},
\end{align}
where the social welfare acts as a normalizer.
As for the compliance rate, we run the experiment several times with different seeds and define the proportion that satisfies the NE constraint with a tolerance of $5\%$ as the Compliance Rate. We also examine revenue and the costs of all platform agents, although they are not the main performance metrics. 

\textbf{Baselines.} We compare our approach to several leading industrial baselines, including multi-agent auto-bidding algorithms, such as MAAB \cite{MAAB} and DCMAB \cite{DCMAB}. We also compare our approach to the state-of-the-art
single-agent auto-bidding algorithms: USCB \cite{USCB}.
See Appendix \ref{app:baselines} for detailed descriptions of these baselines. Furthermore, we compare our approach with an advanced equilibrium selection (ES) algorithm \cite{jothimurugan2022specification}.

\textbf{Practical Implementation.}
To develop a practical algorithm, we employ a Causal Transformer \cite{vaswani2017attention} with 5 Transformer blocks as the neural network backbone.
The adjustment frequency is $1/96$, meaning that we recalibrate the bidding factor using the neural network 96 times during the bidding episode.
The input features of the neural network include the number of impressions arrived, and their click rates, conversion rates, and GMV.

\textbf{Results.}
\cref{table:simulated_exp} gives the results of the experiments under different $\rho$ settings. We see that the NCB framework adapts to the $\rho$ setting: decreasing $\rho$ yields higher Max Exploitability and Social Welfare, whereas other algorithms do not. Importantly, when $10\le \rho\le 50$, the NCB framework achieves both a lower Max Exploitability and a higher Social Welfare compared to all other algorithms with relatively high Compliance Rates ($>65\%$).  

\vspace{-0.5mm}
\section{Conclusions}
\label{sec:conclusions}
We introduced Nash Equilibrium Constrained Bidding (NCB), a scalable framework that unifies game-theoretic equilibrium guarantees with platform-wide optimization for industrial auto-bidding systems. Our approach captures strategic interactions among advertisers while maximizing social welfare, and is supported by rigorous theory and practical algorithms. 
Experiments demonstrate that NCB achieves superior social welfare and equilibrium compliance compared to existing methods. This work bridges the gap between theory and practice in auto-bidding, paving the way for more robust and efficient advertising platforms.

\section*{Impact Statement}

This work advances the field of online advertising by introducing a principled and scalable auto-bidding framework that directly incorporates Nash equilibrium constraints into platform-wide optimization. By rigorously modeling strategic interdependencies among advertisers and aligning individual incentives with ecosystem-level goals, our approach addresses longstanding limitations of both single-agent and heuristic multi-agent methods. Theoretical guarantees and practical algorithms make our solution robust and deployable at industrial scale. We anticipate that this framework will not only improve the stability and efficiency of real-world advertising markets but also inspire further research at the intersection of game theory, optimization, and large-scale machine learning for digital marketplaces.


\bibliography{NCB}

@inproceedings{adikari2015real,
  title={Real time bidding in online digital advertisement},
  author={Adikari, Shalinda and Dutta, Kaushik},
  booktitle={International Conference on Design Science Research in Information Systems},
  pages={19--38},
  year={2015},
  organization={Springer}
}

@article{auto-bidding,
author = {Aggarwal, Gagan and Badanidiyuru, Ashwinkumar and Balseiro, Santiago R. and Bhawalkar, Kshipra and Deng, Yuan and Feng, Zhe and Goel, Gagan and Liaw, Christopher and Lu, Haihao and Mahdian, Mohammad and Mao, Jieming and Mehta, Aranyak and Mirrokni, Vahab and Leme, Renato Paes and Perlroth, Andres and Piliouras, Georgios and Schneider, Jon and Schvartzman, Ariel and Sivan, Balasubramanian and Spendlove, Kelly and Teng, Yifeng and Wang, Di and Zhang, Hanrui and Zhao, Mingfei and Zhu, Wennan and Zuo, Song},
title = {Auto-Bidding and Auctions in Online Advertising: A Survey},
year = {2024},
issue_date = {June 2024},
publisher = {Association for Computing Machinery},
address = {New York, NY, USA},
volume = {22},
number = {1},
url = {https://doi.org/10.1145/3699824.3699838},
doi = {10.1145/3699824.3699838},
journal = {SIGecom Exch.},
month = oct,
pages = {159–183},
numpages = {25}
}

@inproceedings{USCB,
  title={A unified solution to constrained bidding in online display advertising},
  author={He, Yue and Chen, Xiujun and Wu, Di and Pan, Junwei and Tan, Qing and Yu, Chuan and Xu, Jian and Zhu, Xiaoqiang},
  booktitle={Proceedings of the 27th ACM SIGKDD Conference on Knowledge Discovery \& Data Mining},
  pages={2993--3001},
  year={2021}
}

@article{SORL,
  title={Sustainable online reinforcement learning for auto-bidding},
  author={Mou, Zhiyu and Huo, Yusen and Bai, Rongquan and Xie, Mingzhou and Yu, Chuan and Xu, Jian and Zheng, Bo},
  journal={Advances in Neural Information Processing Systems},
  volume={35},
  pages={2651--2663},
  year={2022}
}

@inproceedings{MAAB,
  title={A cooperative-competitive multi-agent framework for auto-bidding in online advertising},
  author={Wen, Chao and Xu, Miao and Zhang, Zhilin and Zheng, Zhenzhe and Wang, Yuhui and Liu, Xiangyu and Rong, Yu and Xie, Dong and Tan, Xiaoyang and Yu, Chuan and others},
  booktitle={Proceedings of the Fifteenth ACM International Conference on Web Search and Data Mining},
  pages={1129--1139},
  year={2022}
}

@inproceedings{DCMAB,
  title={Real-time bidding with multi-agent reinforcement learning in display advertising},
  author={Jin, Junqi and Song, Chengru and Li, Han and Gai, Kun and Wang, Jun and Zhang, Weinan},
  booktitle={Proceedings of the 27th ACM international conference on information and knowledge management},
  pages={2193--2201},
  year={2018}
}

@inproceedings{SAB_1,
  title={Budget constrained bidding by model-free reinforcement learning in display advertising},
  author={Wu, Di and Chen, Xiujun and Yang, Xun and Wang, Hao and Tan, Qing and Zhang, Xiaoxun and Xu, Jian and Gai, Kun},
  booktitle={Proceedings of the 27th ACM International Conference on Information and Knowledge Management},
  pages={1443--1451},
  year={2018}
}

@inproceedings{SAB_2,
  title={Real-time bidding by reinforcement learning in display advertising},
  author={Cai, Han and Ren, Kan and Zhang, Weinan and Malialis, Kleanthis and Wang, Jun and Yu, Yong and Guo, Defeng},
  booktitle={Proceedings of the tenth ACM international conference on web search and data mining},
  pages={661--670},
  year={2017}
}

@article{SAB_3,
  title={Incrementality bidding via reinforcement learning under mixed and delayed rewards},
  author={Badanidiyuru Varadaraja, Ashwinkumar and Feng, Zhe and Li, Tianxi and Xu, Haifeng},
  journal={Advances in Neural Information Processing Systems},
  volume={35},
  pages={2142--2153},
  year={2022}
}

@article{wang2002reinforcement,
  title={Reinforcement learning to play an optimal Nash equilibrium in team Markov games},
  author={Wang, Xiaofeng and Sandholm, Tuomas},
  journal={Advances in neural information processing systems},
  volume={15},
  year={2002}
}

@article{deng2024efficiency,
  title={Efficiency of the first-price auction in the autobidding world},
  author={Deng, Yuan and Mao, Jieming and Mirrokni, Vahab and Zhang, Hanrui and Zuo, Song},
  journal={Advances in Neural Information Processing Systems},
  volume={37},
  pages={139270--139293},
  year={2024}
}

@inproceedings{guo2024generative,
  title={Generative auto-bidding via conditional diffusion modeling},
  author={Guo, Jiayan and Huo, Yusen and Zhang, Zhilin and Wang, Tianyu and Yu, Chuan and Xu, Jian and Zheng, Bo and Zhang, Yan},
  booktitle={Proceedings of the 30th ACM SIGKDD Conference on Knowledge Discovery and Data Mining},
  pages={5038--5049},
  year={2024}
}

@article{milgrom2002envelope,
  title={Envelope theorems for arbitrary choice sets},
  author={Milgrom, Paul and Segal, Ilya},
  journal={Econometrica},
  volume={70},
  number={2},
  pages={583--601},
  year={2002},
  publisher={Wiley Online Library}
}

@article{fischer1992special,
  title={A special Newton-type optimization method},
  author={Fischer, Andreas},
  journal={Optimization},
  volume={24},
  number={3-4},
  pages={269--284},
  year={1992},
  publisher={Taylor \& Francis}
}

@article{conitzer2022multiplicative,
  title={Multiplicative pacing equilibria in auction markets},
  author={Conitzer, Vincent and Kroer, Christian and Sodomka, Eric and Stier-Moses, Nicolas E},
  journal={Operations Research},
  volume={70},
  number={2},
  pages={963--989},
  year={2022},
  publisher={INFORMS}
}

@article{chen2024complexity,
  title={The complexity of pacing for second-price auctions},
  author={Chen, Xi and Kroer, Christian and Kumar, Rachitesh},
  journal={Mathematics of Operations Research},
  volume={49},
  number={4},
  pages={2109--2135},
  year={2024},
  publisher={INFORMS}
}

@article{conitzer2022pacing,
  title={Pacing equilibrium in first price auction markets},
  author={Conitzer, Vincent and Kroer, Christian and Panigrahi, Debmalya and Schrijvers, Okke and Stier-Moses, Nicolas E and Sodomka, Eric and Wilkens, Christopher A},
  journal={Management Science},
  volume={68},
  number={12},
  pages={8515--8535},
  year={2022},
  publisher={INFORMS}
}

@inproceedings{lucier2024autobidders,
  title={Autobidders with budget and roi constraints: Efficiency, regret, and pacing dynamics},
  author={Lucier, Brendan and Pattathil, Sarath and Slivkins, Aleksandrs and Zhang, Mengxiao},
  booktitle={The Thirty Seventh Annual Conference on Learning Theory},
  pages={3642--3643},
  year={2024},
  organization={PMLR}
}

@inproceedings{liu2023auto,
  title={Auto-bidding with Budget and ROI Constrained Buyers.},
  author={Liu, Xiaodong and Shen, Weiran},
  booktitle={IJCAI},
  pages={2817--2825},
  year={2023}
}

@book{horn2012matrix,
  title={Matrix analysis},
  author={Horn, Roger A and Johnson, Charles R},
  year={2012},
  publisher={Cambridge university press}
}

@article{bertsekas1997nonlinear,
  title={Nonlinear programming},
  author={Bertsekas, Dimitri P},
  journal={Journal of the Operational Research Society},
  volume={48},
  number={3},
  pages={334--334},
  year={1997},
  publisher={Taylor \& Francis}
}

@inproceedings{aggarwal2019autobidding,
  title={Autobidding with constraints},
  author={Aggarwal, Gagan and Badanidiyuru, Ashwinkumar and Mehta, Aranyak},
  booktitle={International Conference on Web and Internet Economics},
  pages={17--30},
  year={2019},
  organization={Springer}
}

@article{su2024auctionnet,
  title={AuctionNet: A Novel Benchmark for Decision-Making in Large-Scale Games},
  author={Su, Kefan and Huo, Yusen and Zhang, Zhilin and Dou, Shuai and Yu, Chuan and Xu, Jian and Lu, Zongqing and Zheng, Bo},
  journal={arXiv preprint arXiv:2412.10798},
  year={2024}
}

@inproceedings{jothimurugan2022specification,
  title={Specification-guided learning of nash equilibria with high social welfare},
  author={Jothimurugan, Kishor and Bansal, Suguman and Bastani, Osbert and Alur, Rajeev},
  booktitle={International Conference on Computer Aided Verification},
  pages={343--363},
  year={2022},
  organization={Springer}
}

@article{MADDPG,
  title={Multi-agent actor-critic for mixed cooperative-competitive environments},
  author={Lowe, Ryan and Wu, Yi I and Tamar, Aviv and Harb, Jean and Pieter Abbeel, OpenAI and Mordatch, Igor},
  journal={Advances in neural information processing systems},
  volume={30},
  year={2017}
}

@article{vaswani2017attention,
  title={Attention is all you need},
  author={Vaswani, Ashish and Shazeer, Noam and Parmar, Niki and Uszkoreit, Jakob and Jones, Llion and Gomez, Aidan N and Kaiser, {\L}ukasz and Polosukhin, Illia},
  journal={Advances in neural information processing systems},
  volume={30},
  year={2017}
}

@article{mou2025enhancing,
  title={Enhancing Generative Auto-bidding with Offline Reward Evaluation and Policy Search},
  author={Mou, Zhiyu and Lv, Yiqin and Xu, Miao and Wang, Qi and Mao, Yixiu and Ye, Qichen and Li, Chao and Bai, Rongquan and Yu, Chuan and Xu, Jian and others},
  journal={arXiv preprint arXiv:2509.15927},
  year={2025}
}

@inproceedings{paes2024complex,
  title={Complex dynamics in autobidding systems},
  author={Paes Leme, Renato and Piliouras, Georgios and Schneider, Jon and Spendlove, Kelly and Zuo, Song},
  booktitle={Proceedings of the 25th ACM Conference on Economics and Computation},
  pages={75--100},
  year={2024}
}

@article{balseiro2019learning,
  title={Learning in repeated auctions with budgets: Regret minimization and equilibrium},
  author={Balseiro, Santiago R and Gur, Yonatan},
  journal={Management Science},
  volume={65},
  number={9},
  pages={3952--3968},
  year={2019},
  publisher={INFORMS}
}

@article{nash1950equilibrium,
  title={Equilibrium points in n-person games},
  author={Nash Jr, John F},
  journal={Proceedings of the national academy of sciences},
  volume={36},
  number={1},
  pages={48--49},
  year={1950},
  publisher={national academy of sciences}
}

@article{matrix_game,
  title={Equilibrium identification and selection in finite games},
  author={Cr{\"o}nert, Tobias and Minner, Stefan},
  journal={Operations Research},
  volume={72},
  number={2},
  pages={816--831},
  year={2024},
  publisher={INFORMS}
}

@article{ne_selection_matrix_game,
  title={An approach to equilibrium selection},
  author={Matsui, Akihiko and Matsuyama, Kiminori},
  journal={Journal of Economic Theory},
  volume={65},
  number={2},
  pages={415--434},
  year={1995},
  publisher={Elsevier}
}

@inproceedings{ne_selection_matrix,
  title={Approximate Nash equilibria with near optimal social welfare},
  author={Czumaj, Artur and Fasoulakis, Michail and Jurdzinski, Marcin},
  booktitle={Proceedings of the 24th International Conference on Artificial Intelligence},
  pages={504--510},
  year={2015}
}

@article{ne_selection_matrix_2,
  title={How hard is it to approximate the best Nash equilibrium?},
  author={Hazan, Elad and Krauthgamer, Robert},
  journal={SIAM Journal on Computing},
  volume={40},
  number={1},
  pages={79--91},
  year={2011},
  publisher={SIAM}
}

@inproceedings{ne_selection_markov_game,
  title={Specification-guided learning of nash equilibria with high social welfare},
  author={Jothimurugan, Kishor and Bansal, Suguman and Bastani, Osbert and Alur, Rajeev},
  booktitle={International Conference on Computer Aided Verification},
  pages={343--363},
  year={2022},
  organization={Springer}
}

@inproceedings{ne_selection_2_value,
  title={Maximizing Nash social welfare in 2-value instances},
  author={Akrami, Hannaneh and Chaudhury, Bhaskar Ray and Hoefer, Martin and Mehlhorn, Kurt and Schmalhofer, Marco and Shahkarami, Golnoosh and Varricchio, Giovanna and Vermande, Quentin and van Wijland, Ernest},
  booktitle={Proceedings of the AAAI Conference on Artificial Intelligence},
  volume={36},
  number={5},
  pages={4760--4767},
  year={2022}
}

@article{NES,
  title={Reinforcement learning to play an optimal Nash equilibrium in team Markov games},
  author={Wang, Xiaofeng and Sandholm, Tuomas},
  journal={Advances in neural information processing systems},
  volume={15},
  year={2002}
}

@misc{roughgarden2009algorithmic,
  title={Algorithmic game theory},
  author={Roughgarden, Tim},
  year={2009},
  publisher={Communica}
}

@inproceedings{chiappa2025auto,
  title={Auto-bidding in real-time auctions via Oracle Imitation Learning},
  author={Chiappa, Alberto Silvio and Gangopadhyay, Briti and Wang, Zhao and Takamatsu, Shingo},
  booktitle={Proceedings of the 31st ACM SIGKDD Conference on Knowledge Discovery and Data Mining V. 2},
  pages={345--356},
  year={2025}
}
\bibliographystyle{icml2026}

\newpage
\appendix
\onecolumn

\section{Related Works}
\label{app:related_work}

\subsection{large-scale industrial auto-bidding Algorithms}
\label{app:related_work_auto_bidding}
\subsubsection{Single-Agent Auto-bidding Algorithms}


Most existing large-scale auto-bidding research focuses on Single-Agent auto-Bidding (SAB), aiming to maximize the expected return for an individual advertiser under specific constraints \cite{USCB, SORL, SAB_1, SAB_2, SAB_3, guo2024generative, mou2025enhancing, chiappa2025auto}. 
Reinforcement Learning (RL) has been the cornerstone of these approaches. For example, USCB \cite{USCB} uses the Deep Deterministic Policy Gradient (DDPG) algorithm to learn policies within a simulator, while V-CQL \cite{SORL} employs an offline RL framework to train policies directly from fixed offline datasets. More recently, generative models have introduced a new paradigm to the field. Specifically, AIGB \cite{guo2024generative} models auto-bidding as a trajectory generation task, approximating the conditional distribution of offline data. Based on this, AIGB-Pearl \cite{mou2025enhancing} integrates generative planning with a KL-Lipschitz-constrained score-maximization scheme, enabling safe and efficient exploration beyond the limitations of fixed offline datasets. 
Despite these advances, SAB algorithms are theoretically limited in scenarios that require simultaneous control of multiple advertisers. As discussed in Section \ref{sec:intro}, it completely neglects mutual influence among agents, resulting in weaker equilibrium guarantees and suboptimal advertising outcomes. 

\subsubsection{Multi-Agent Auto-bidding Algorithms}

Other algorithms investigate multi-agent approaches for auto-bidding \cite{MAAB, DCMAB}, they typically cluster agents into a few coarse groups and model only inter-cluster interactions, failing to capture intra-cluster strategic dependencies in a rigorous or game-theoretically consistent manner, as shown in Fig.\ref{fig:auto_bidding_process}(b). This often results in suboptimal bidding policies and distorted equilibrium outcomes within groups. Moreover, these approaches lack a principled problem formulation that explicitly and coherently captures the coupling between advertiser-level utilities and platform-wide metrics. Consequently, they often resort to heuristic methods without provable guarantees for equilibrium convergence or system-wide performance.

In summary, a principled and scalable auto-bidding framework that jointly accounts for fine-grained strategic interdependencies among agents and the platform’s ecosystem-level goals remains an open challenge.

\subsection{Equilibrium in auto-bidding problem}
\label{app:related_work_equ_in_autobidding}
With the widespread adoption of auto-bidding among advertisers, system-wide equilibrium has emerged as a critical factor in governing global market dynamics and advertiser outcomes.
This line of work investigates pacing equilibria \cite{conitzer2022multiplicative, conitzer2022pacing}, a state where no advertiser can improve their value by unilaterally deviating from their pacing strategy. While foundational studies establish the existence, and potential multiplicity, of such Nash equilibria \cite{nash1950equilibrium}, they also demonstrate that computing an exact equilibrium is PPAD-complete \cite{conitzer2022multiplicative, chen2024complexity}, rendering it computationally intractable for large-scale systems. 
To address this, \cite{conitzer2022multiplicative} proposed a mixed-integer programming (MIP) formulation to compute equilibrium bids; however, its exponential complexity limits its applicability to small-scale scenarios. 
Other studies \cite{balseiro2019learning, lucier2024autobidders, liu2023auto} focus on the convergence of system dynamics under specific regularity conditions, and provide theoretical lower bounds for the social welfare or the PoA (price of anarchy \cite{roughgarden2009algorithmic}).
Despite these diverse theoretical contributions, a fundamental gap remains across existing literature. First, current methods generally lack the mechanism to converge to a preferred equilibrium (e.g. one that maximizes social welfare or platform revenue—among the potentially many that exist). Second, these approaches typically fail to provide the scalability required for real-world industrial systems involving millions of advertisers.


\subsection{Equilibrium Selection Algorithms}
\label{app:es}

Inherently, the NCB problem falls into the category of the ES problem, whose objective is to find the best $\epsilon$-NE based on certain criteria among all available ones.
There are some NE selection methods for stateless matrix games \cite{matrix_game, ne_selection_matrix_game, ne_selection_matrix, ne_selection_matrix_2, ne_selection_2_value} that are relatively simpler compared to POMG considered in this paper and cannot be applied directly to it. 
In addition, \cite{NES} studies maximize social welfare under $\epsilon$-NE in a team Markov game where all agents share the payoff and have no conflicts. Because agents can exhibit significant interest conflicts in the NCB problem and do not share rewards, the algorithm in \cite{NES} is not suitable for NCB.
Recently, there have been a few works discussing the NE selection in general Markov games \cite{ne_selection_markov_game}. However, they are basically based on searching the joint policy with the largest social welfare in a candidate set of $\epsilon$-NEs or searching the $\epsilon$-NE in a candidate set of joint policies with large social welfare. For example, the algorithm in \cite{ne_selection_markov_game} first enumerates all the feasible optimal joint policies in a prioritized rank and checks from the one with the largest social welfare if it satisfies the $\epsilon$-NE by adding extra strategies.


\section{Proof of Lemma \ref{lemma:opt}}
\label{app:proof_lemma}
\renewcommand{\thetheorem}{3.1}
\begin{lemma}[Unilateral Best Response Condition]
    The optimal bidding factor $\alpha_i$ for agent $i$'s best-response problem in Eq.~\eqref{equ:single_agent_problem} either exhausts agent $i$'s budget or saturates at the upper bound $A$, whichever occurs first, i.e., 
    \begin{align}
        C_i(\alpha_iv_i;b_{-i})=\min(B_i,C_i(Av_i;b_{-i})),
    \end{align}
    where $v_i\triangleq[v_{i,1},v_{i,2},\cdots,v_{i,K}]$.
\end{lemma}
\begin{proof}
From the definition of $R_i(b_i;b_{-i})$ and $C_i(b_i;b_{-i})$ in Eq. \eqref{equ:single_agent_problem}, it can be known that both the return and the cost functions are summations of a series of softmax functions, which are monotonically increasing with respect to the bid on each impression $b_{i,k}$. As the optimal bid takes the form $b_{i,k}=\alpha_i v_{i,k}$ as given in Eq. \eqref{equ:bid_stru}, it can be known that both the return and the cost functions are monotonically increasing with respect to $\alpha_i$. Therefore, to obtain as much return as possible for agent $i$, $\alpha_i$ should be as large as possible until the budget $B_i$ runs out or $\alpha_i$ is saturated. 
Specifically, there are two cases:
\begin{itemize}
    \item If the cost of the maximum bidding factor exceeds the budget, i.e., $C_i(Av_i;b_{-i})> B_i$, then the cost should be bounded by the budget $B_i$. Specifically, as $C_i(\alpha_iv_i;b_{-i})$ is a continuous function with respect to $\alpha_i$, there exists a threshold bidding factor such that the budget is exactly exhausted, i.e., $C_i(\alpha_iv_i;b_{-i})=B_i$.
    \item If the cost of the maximum bidding factor is no more than the budget, i.e., $C_i(Av_i;b_{-i})\le B_i$, then the bidding factor $\alpha_i$ should exactly be $A$ to maximize the return, and the cost should be $C_i(Av_i;b_{-i})$.
\end{itemize}
Summarizing these two cases, we can have the optimal condition for $\alpha_i$ being optimal:
\begin{align}
    C_i(\alpha_iv_i)=\min (B_i, C_i(Av_i;b_{-i})).
\end{align}
This completes the proof.
\end{proof}

\section{Illustrative Example}
\label{app:example}


\begin{table}[t]
    \centering
    \begin{tabular}{c|c|c|c|c|c|c|c|c|c|c}
   \hline
       Impression Values  & $v_{i,1}$&$v_{i,2}$&$v_{i,3}$&$v_{i,4}$&$v_{i,5}$&$v_{i,6}$&$v_{i,7}$&$v_{i,8}$&$v_{i,9}$&$v_{i,10}$ \\
       \hline
        Agent 1 &3.645&2.806&0.624&1.987&3.906& 2.554&0.913&4.921 &4.776&4.267
        \\
\hline
        Agent 2& 0.963&4.853&1.174&0.131&4.730& 4.608&1.469&2.553 &1.977&0.831
        \\
        \hline
        Agent 3& 1.540&2.147&4.150&2.811&4.154& 4.984&1.662&0.746 &3.869&0.455
  \\
         \hline
    \end{tabular}
    \caption{Impression values of each agent in the illustrative example on Nash Equilibrium bidding.}
    \label{tab:example}
\end{table}



        


\begin{table}[t]
    \centering
      \begin{tabular}{c|c|c|c|c}
    \hline
        Nash Equilibrium 1 & Agent 1& Agent 2 & Agent 3 & Total\\
       \hline
        Budget & 7.254 & 9.561 & 0.731 &17.546\\
        \hline
        $\alpha^*$ & 0.664 & 1.290&0.361&-\\
        \hline
        Cost & 7.253 & 9.561&0.731&17.545 (Revenue)\\
\hline
Value & 16.712&17.766&1.982&\textbf{36.462} (Social Welfare)
\\
\hline
    \end{tabular}
    
\vspace{1em}

    \begin{tabular}{c|c|c|c|c}
    \hline
        Nash Equilibrium 2 & Agent 1& Agent 2 & Agent 3 & Total\\
       \hline
        Budget & 7.254 & 9.561 & 0.731 &17.546\\
        \hline
        $\alpha^*$ & 1.015 & 0.856&0.262&-\\
        \hline
        Cost & 7.253 & 9.561&0.731&17.545 (Revenue)\\
\hline
Value & 20.625&14.699&3.017&\textbf{38.368} (Social Welfare)
\\
\hline  
    \end{tabular}
    \caption{Two Nash Equilibria of the illustrative example. The budgets of agents are $7.254$, $9.561$, and $0.731$, respectively.}
    \label{tab:ne}
\end{table}

\begin{figure}[htbp]
\centering
\begin{subfigure}[b]{0.48\textwidth}
    \centering
    \includegraphics[width=\linewidth]{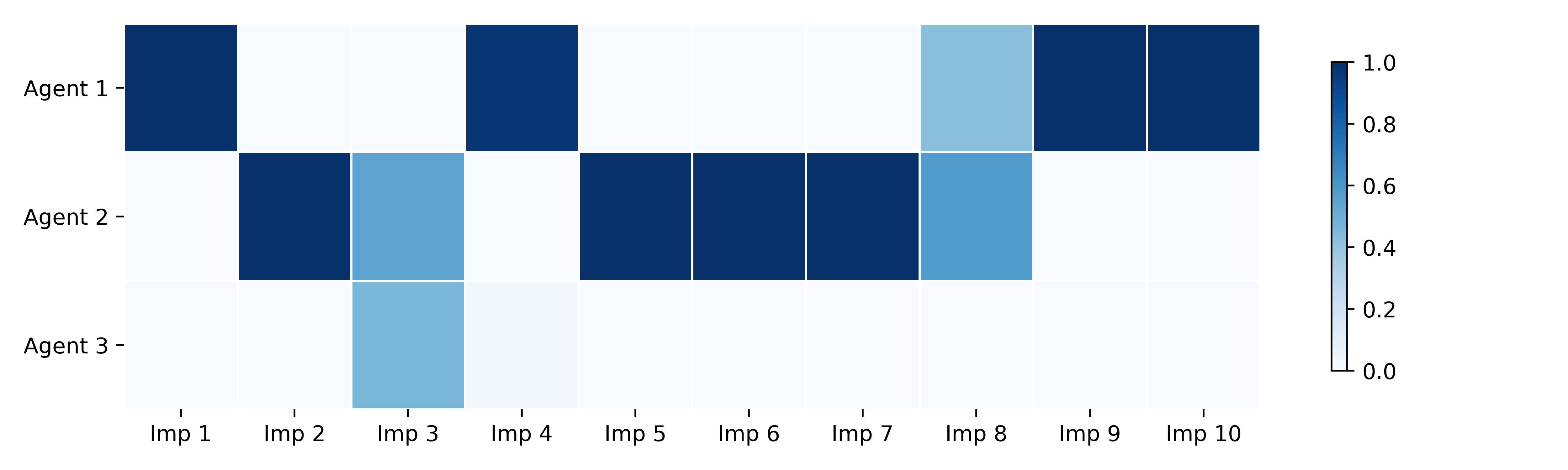}
    \caption{Winning probability in Nash Equilibrium 1.}
    \label{fig:wb_ne1}
\end{subfigure}
\hfill
\begin{subfigure}[b]{0.48\textwidth}
    \centering
    \includegraphics[width=\linewidth]{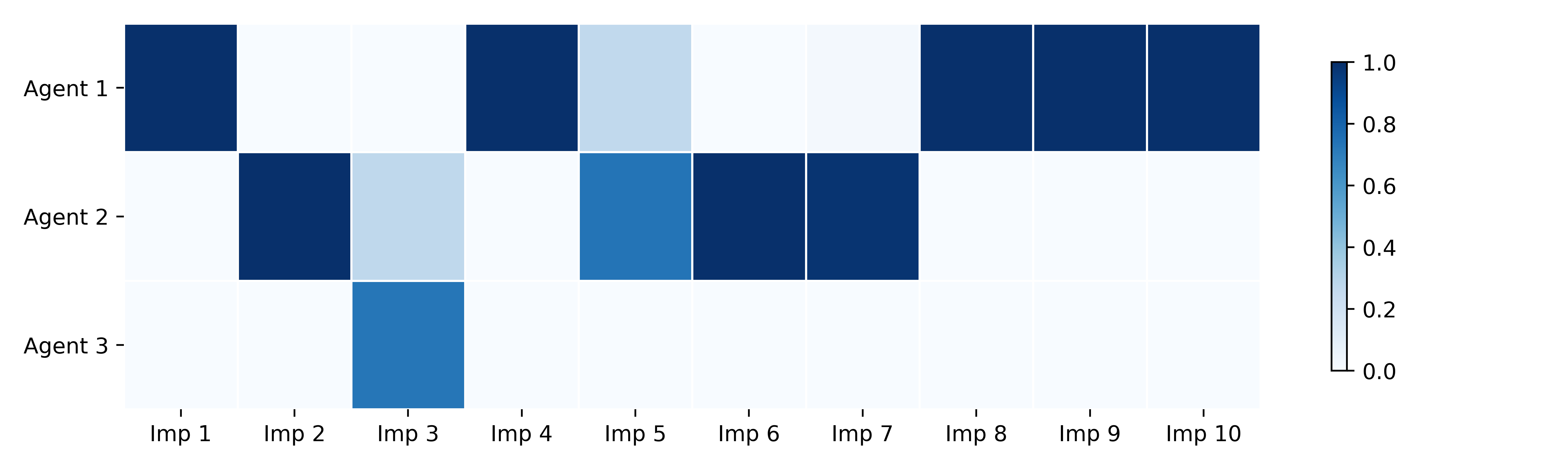}
    \caption{Winning probability in Nash Equilibrium 2.}
    \label{fig:wb_ne2}
\end{subfigure}
\caption{Winning probabilities of each agent for every impression in two Nash Equilibria.}
\label{fig:wb_ne}
\end{figure}

We present an illustrative example of the Nash Equilibrium-Constrained Bidding. Consider a setting with $N=3$ agents competing for $K=10$ impressions. The agents' budgets are $7.254$, $9.561$, and $0.731$, respectively. Their respective impression valuations are given in Table \ref{tab:example}.  
As shown in Table. \ref{tab:ne}, there exist two distinct NE points. 
At each equilibrium, every agent exhausts its budget, and any attempt to increase $\alpha$ to improve its valuation would violate the budget constraint. 
However, these two equilibria exhibit a notable difference in social welfare.
Specifically, the relative difference in social welfare between the two Nash equilibria is $\textbf{5.2\%}$ (=$|38.368-36.462|/36.462$).
The winning probabilities of each agent for every impression in two equilibria shown in Fig. \ref{fig:wb_ne}.
We can see that the winning probabilities of agents on Impression 3, 5, and 8 are significantly different at two NE points. 
Especially, Agent 2 tends to win both Impression 3 and Impression 8 at NE point 1 ($p_{2,3}=0.539, ,p_{2,8}=0.576$), while Agent 3 tends to win Impression 3 and Agent 1 tends to win Impression 8 at NE point 2 ($p_{3,3}=0.727,p_{1,8}=1.0$).

\begin{figure}[t]
    \centering
\includegraphics[width=0.5\linewidth]{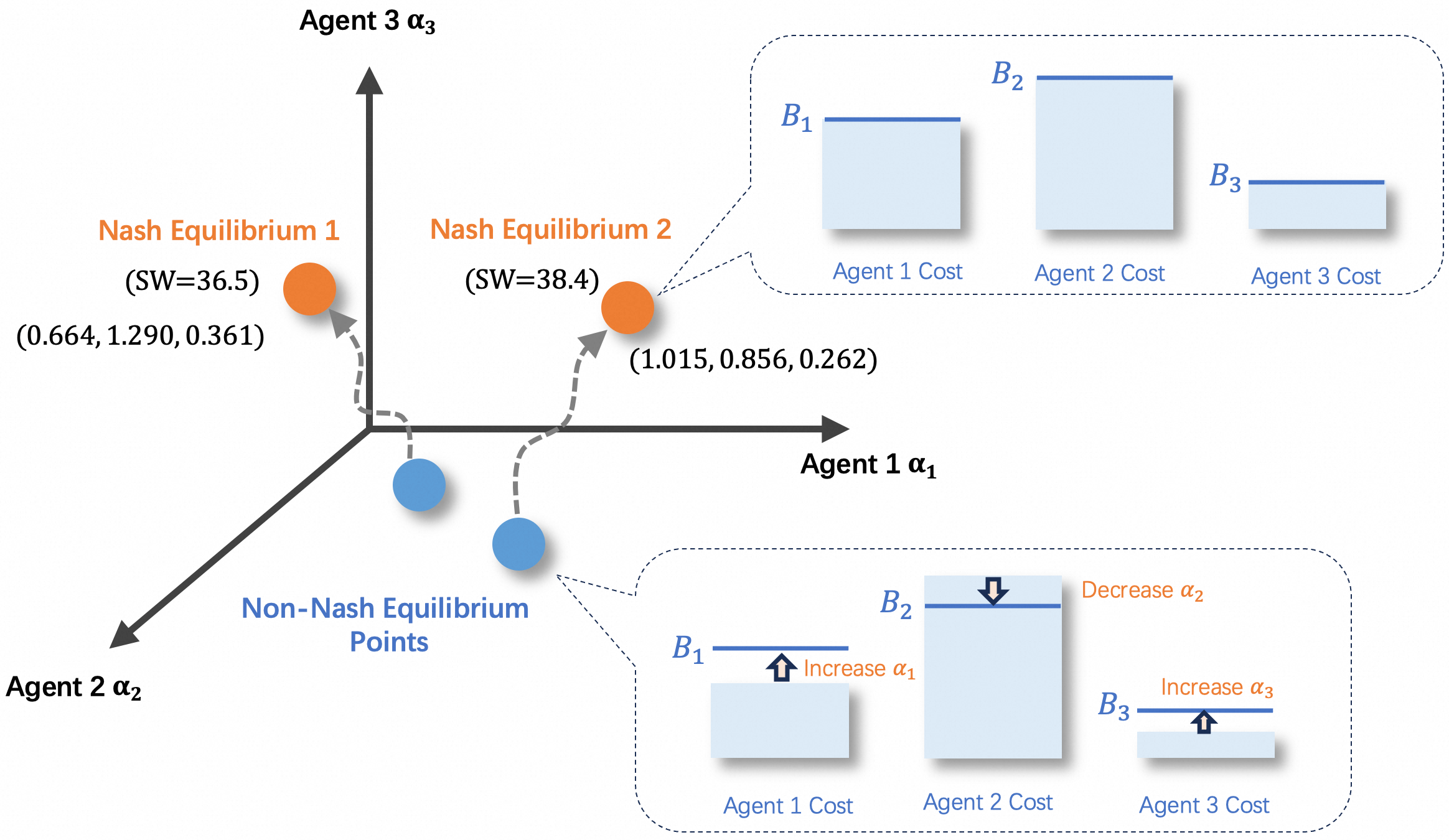}
    \caption{The strategic dynamics in Example \ref{example}. Two distinct NE points exist, both of which are characterized by full budget exhaustion across all agents. Nevertheless, their respective social welfare outcomes exhibit a $5.2\%$ relative disparity.}
    \label{fig:example_game2}
\end{figure}

\section{Proof of Theorem \ref{thm:equivalent_ne}}
\label{app:proof_thm_equ_ne}
\renewcommand{\thetheorem}{4.1}
\begin{theorem}[Equivalent NE Constraint]
When $\epsilon=0$, $h_i(\alpha)=0,\forall i$ is equivalent to the NE constraints in Eq.~\eqref{equ:ncb_alpha}.
\end{theorem}
\begin{proof}
Recall that when $\epsilon=0$, if $h_i(\alpha)=0$, then it indicates that:
\begin{align}
\label{equ:h_i_0}
    B_i-C_i(\alpha)+A-\alpha_i-\sqrt{(B_i-C_i(\alpha))^2+(A-\alpha_i)^2}=0.
\end{align}
The constraints in Eq.~\eqref{equ:ncb_alpha} are, $\forall i$ \footnote{Here we note that as $B_i>0$ and $\alpha_i$ is the optimal bidding factor of agent $i$, $\alpha_i$ is sure to be non-negative, i.e., $\alpha_i\ge 0$.
Therefore, we use the condition $\alpha_i\le A$ instead of $\alpha_i\in[0,A]$ in Eq. \eqref{equ:org_ne}.}:
\begin{align}
\label{equ:org_ne}
    C_i(\alpha_i;\alpha_{-i})=\min(B_i, C_i(A;\alpha_{-i})),\quad \alpha_i\le A.
\end{align}
Specifically, when $h_i(\alpha)=0$, it indicates that 
\begin{align}
&\qquad h_i(\alpha)=0\notag\\
&\Leftrightarrow
     B_i-C_i(\alpha)+A-\alpha_i=\sqrt{(B_i-C_i(\alpha))^2+(A-\alpha_i)^2}.\notag\\
    & \Leftrightarrow \bigg(B_i-C_i(\alpha)+A-\alpha_i\bigg)^2=(B_i-C_i(\alpha))^2+(A-\alpha_i)^2 \text{ and } B_i-C_i(\alpha)+A-\alpha_i\ge 0\notag\\
    & \Leftrightarrow (B_i-C_i(\alpha))(A-\alpha_i)=0 \text{ and } B_i-C_i(\alpha)+A-\alpha_i\ge 0.
\end{align} 
This is equivalent to:
\begin{align}
\label{equ:equi_h_i}
    \bigg(B_i=C_i(\alpha),\quad \alpha_i\le A\bigg) \quad\text{or} \quad \bigg(A=\alpha_i, \quad C_i(A;\alpha_{-i})\le B_i\bigg),
\end{align}
holds at the same time.
Moreover, 
Eq. \eqref{equ:org_ne} is equivalent to
\begin{align}
\label{equ:equ_tmp}
  & B_i\le C_i(A;\alpha_{-i}), \quad C_i(\alpha)=B_i,\quad\alpha_i\le A\notag\\
   \text{or}\quad&C_i(A;\alpha)\le B_i,\quad C_i(\alpha_i;\alpha_{-i})=C_i(A;\alpha_{-i}),\quad\alpha_i\le A
\end{align}
holds at the same time.
As $C_i(\alpha_i;\alpha_{-i})$ monotonically increases with $\alpha_i$, $C_i(\alpha_i;\alpha_{-i})=C_i(A;\alpha_{-i})$ is equivalent to $\alpha_i=A$, and when $\alpha_i\le A$, it holds that $C_i(\alpha_i;\alpha_{-i})\le C_i(A;\alpha_{-i})$.
Therefore, the conditions in Eq. \eqref{equ:equ_tmp} are equivalent to:
\begin{align}
     \bigg(B_i=C_i(\alpha),\quad \alpha_i\le A\bigg) \quad\text{or} \quad \bigg(A=\alpha_i, \quad C_i(A;\alpha_{-i})\le B_i\bigg),
\end{align}
holds at the same time. This is equivalent to Eq. ~\eqref{equ:equi_h_i}.
Therefore, Eq. \eqref{equ:h_i_0} is equivalent to Eq. \eqref{equ:org_ne}, which completes the proof.
\end{proof}

\section{Theoretical Gradient Derivations}
\label{app:proof_gradients}

Recall that the augmented Lagrangian function for the NCB problem in Eq. \eqref{equ:ncb_alpha_simplify} is:
\begin{align}
     L_\rho(\alpha, \lambda)
    =\sum_{i\in [N]}\bigg[R_i(\alpha)+\lambda_i h_i(\alpha)
    -\frac{\rho}{2} h_i^2(\alpha)\bigg],
\end{align}
where $R_i(\alpha)$ and $h_i(\alpha)$ are the return and the cost of agent $i$, i.e., 
\begin{align}
R_i(\alpha)=\sum_{k\in[K]}p_{i,k}(\alpha)\times v_{i,k},\quad C_i(\alpha)=\sum_{k\in[K]}p_{i,k}(\alpha)\times \underbrace{\sum_{j\neq i}p^{-i}_{j,k}(\alpha)\times\alpha_{j}v_{j,k}}_{= m_{i,k}(\alpha_{-i})}.
\end{align}
and $h_i(\alpha)$ is the constraint for agent $i$, i.e.,
\begin{align}
    h_i(\alpha)=B_i-C_i(\alpha_i;\alpha_{-i})+A-\alpha_i-\sqrt{(B_i-C_i(\alpha_i;\alpha_{-i}))^2+(A-\alpha_i)^2+\epsilon},
\end{align}
and the winning probability $p_{i,k}(\alpha)$ is expressed as:
\begin{align}
    p_{i,k}(\alpha) = \frac{\exp{(\alpha_iv_{i,k}/\tau)}}{\sum_j \exp{(\alpha_jv_{j,k}/\tau)}}.
\end{align}
Before diving into the detailed derivations, we derive the gradients of some critical terms.

\textbf{(1) The gradient of winning probability: $\frac{\partial p_{i,k}(\alpha)}{\partial \alpha_j}$}.
\begin{align}
\label{equ:nabla_p}
    \frac{\partial p_{i,k}(\alpha)}{\partial \alpha_j}&=p_{i,k}\frac{v_{i,k}}{\tau}\delta_{ij}-p_{i,k}\frac{\exp{(\alpha_j v_{j,k}/\tau)}}{\sum_j \exp(\alpha_jv_{j,k}/\tau)}\frac{v_{j,k}}{\tau}\notag\\
    &=\frac{p_{i,k}}{\tau}\bigg[v_{i,k}\delta_{ij}-p_{j,k}v_{j,k}\bigg]\notag\\
    &=\frac{p_{i,k}v_{j,k}}{\tau}(\delta_{ij}-p_{j,k}).
\end{align}
where $\delta_{ij}\triangleq\mathbf{1}\{i=j\}$ is the Kronecker delta.

\textbf{(2) The gradient of the market price: $\frac{\partial m_{i,k}(\alpha_{-i})}{\partial \alpha_j} $.}
For the gradient of $m_{i,k}(\alpha_{-i})$ with respect to $\alpha_i$, it holds that:
\begin{align}
    \frac{\partial m_{i,k}(\alpha_{-i})}{\partial \alpha_i}=0.
\end{align}
For the gradient of $m_{i,k}(\alpha_{-i})$ with respect to $\alpha_j, j\neq i$, it holds that:
\begin{align}
     \frac{\partial m_{i,k}(\alpha_{-i})}{\partial \alpha_j}&=\frac{\partial}{\partial\alpha_j}\sum_{u\neq i}p^{-i}_{u,k} \alpha_uv_{u,k}\notag\\
     &=\sum_{u\neq i}\bigg[\alpha_uv_{u,k}\frac{\partial p^{-i}_{u,k}}{\partial \alpha_j}+p^{-i}_{u,k}v_{u,k}\delta_{uj}\bigg]\notag\\
     &=\sum_{u\neq i}\bigg[\alpha_uv_{u,k}\frac{p^{-i}_{u,k}v_{j,k}}{\tau}(\delta_{uj}-p^{-i}_{j,k})+p^{-i}_{u,k}v_{u,k}\delta_{uj}\bigg]\notag\\
     &=\sum_{u\neq i}\bigg[\delta_{uj}p^{-i}_{u,k}v_{u,k}\bigg(1+\frac{\alpha_uv_{j,k}}{\tau}\bigg)-p^{-i}_{j,k}\frac{v_{j,k}}{\tau}p^{-i}_{u,k}v_{u,k}\alpha_u\bigg]\notag\\
     &=p^{-i}_{j,k}v_{j,k}\bigg(1+\frac{\alpha_j v_{j,k}}{\tau}\bigg)-p^{-i}_{j,k}\frac{v_{j,k}}{\tau}\underbrace{\sum_{u\neq i}p^{-i}_{u,k}v_{u,k}\alpha_u}_{=m_{i,k}}\notag\\
     &=p^{-i}_{j,k}v_{j,k}\bigg[1+\frac{\alpha_j v_{j,k}-m_{i,k}}{\tau}\bigg]
\end{align}
Combining the above two results, we have:
\begin{align}
\label{equ:partial_m}
    \frac{\partial m_{i,k}(\alpha_{-i})}{\partial \alpha_j}=p^{-i}_{j,k}v_{j,k}\bigg[1+\frac{\alpha_j v_{j,k}-m_{i,k}}{\tau}\bigg](1-\delta_{ij}),\quad\forall j.
\end{align}


\textbf{(3) The gradient of the return: $\frac{\partial R_i(\alpha)}{\partial \alpha_j}$.} With the results in Eq.~\eqref{equ:nabla_p}, we have:
\begin{align}
    \frac{\partial R_i(\alpha)}{\partial \alpha_j} = \sum_{k\in [K]} v_{i,k}\frac{\partial p_{i,k}(\alpha)}{\partial \alpha_j}=\sum_{k\in [K]} v_{i,k}\frac{p_{i,k}v_{j,k}}{\tau}(\delta_{ij}-p_{j,k}),\quad \forall j.
\end{align}

\textbf{(4) The gradient of the cost: $\frac{\partial C_i(\alpha)}{\partial \alpha_j}$.}
\begin{align}
\label{equ:nabla_C}
    \frac{\partial C_i(\alpha)}{\partial \alpha_j}& =\sum_{k\in [K]}\bigg[p_{i,k}(\alpha) \frac{\partial m_{i,k}(\alpha_{-i})}{\partial \alpha_j} + \frac{\partial p_{i,k}(\alpha)}{\partial \alpha_j}m_{i,k}(\alpha_{-i})\bigg]\notag\\
    &=\sum_{k\in [K]}\bigg[p_{i,k}p^{-i}_{j,k}v_{j,k}\bigg[1+\frac{\alpha_j v_{j,k}-m_{i,k}}{\tau}\bigg](1-\delta_{ij})+m_{i,k}\frac{p_{i,k}v_{j,k}}{\tau}(\delta_{ij}-p_{j,k})\bigg].
\end{align}

\textbf{(5) The gradient of $h_i(\alpha)$: $\frac{\partial h_i(\alpha)}{\partial \alpha_j}$.}
\begin{align}
\label{equ:nabla_h}
    \frac{\partial h_i(\alpha)}{\partial \alpha_j}&=-\frac{\partial C_i(\alpha)}{\partial\alpha_j}-\delta_{ij}+\frac{(B_i-C_i(\alpha))\frac{\partial C_i(\alpha)}{\partial\alpha_j}+(A-\alpha_i)\delta_{ij}}{\sqrt{(B_i-C_i(\alpha))^2+(A-\alpha_i)^2+\epsilon}}\notag\\
    &=-\bigg(1-\frac{B_i-C_i(\alpha)}{Z_i}\bigg)\frac{\partial C_i(\alpha)}{\partial\alpha_j}-\bigg(1-\frac{A-\alpha_i}{Z_i}\bigg)\delta_{ij},
\end{align}
where $Z_i\triangleq \sqrt{(B_i-C_i(\alpha))^2+(A-\alpha_i)^2+\epsilon}$.

Equipped with the above results, we can derive the gradient of $L_\rho$ with respect to $\alpha$ and $\lambda$.

\subsection{Gradient of $L_\rho(\alpha,\lambda)$ with respect to $\alpha$}

The gradient of $L_\rho(\alpha,\lambda)$ with respect to $\alpha$ is:
\begin{align}
    \nabla_\alpha L_\rho(\alpha,\lambda)=\bigg[\frac{\partial L_\rho (\alpha,\lambda)}{\partial \alpha_1}, \frac{\partial L_\rho (\alpha,\lambda)}{\partial \alpha_2},\cdots, \frac{\partial L_\rho (\alpha,\lambda)}{\partial \alpha_N}\bigg]^T
\end{align}
where $\forall j$, we have

\begin{align}
    \frac{\partial L_\rho (\alpha,\lambda)}{\partial \alpha_j}&=\sum_{i\in [N]}\bigg[\frac{\partial R_i(\alpha)}{\partial \alpha_j}+(\lambda_i-\rho h_i(\alpha))\frac{\partial h_i(\alpha)}{\partial \alpha_j}\bigg]\notag\\
    &=\sum_{i\in [N]}\sum_{k\in [K]}
    v_{i,k}\frac{p_{i,k}v_{j,k}}{\tau}(\delta_{ij}-p_{j,k})+\sum_{i\in[N]}(\lambda_i-\rho h_i(\alpha))\frac{\partial h_i(\alpha)}{\partial \alpha_j}\notag\\
    &=\sum_{k\in [K]}\frac{p_{j,k}v_{j,k}}{\tau}(v_{j,k}-\bar{v}_k)+\sum_{i\in [N]}\bigg(\lambda_i-\rho h_i(\alpha)\bigg)\bigg[ -\bigg(1+\frac{B_i+C_i(\alpha)}{Z_i}\bigg)\frac{\partial C_i(\alpha)}{\partial\alpha_j}-\bigg(1+\frac{A-\alpha_i}{Z_i}\bigg)\delta_{ij}\bigg],
\end{align}
where $\frac{\partial C_i(\alpha)}{\partial \alpha_j}$ is given in Eq.~\eqref{equ:nabla_C}.

\subsection{Gradient of $L_\rho(\alpha,\lambda)$ with respect to $\lambda$}

The gradient of $L_\rho(\alpha,\lambda)$ with respect to $\lambda$ is:
\begin{align}
    \nabla_\lambda L_\rho(\alpha,\lambda)=\bigg[\frac{\partial L_\rho (\alpha,\lambda)}{\partial \lambda_1}, \frac{\partial L_\rho (\alpha,\lambda)}{\partial \lambda_2},\cdots, \frac{\partial L_\rho (\alpha,\lambda)}{\partial \lambda_N}\bigg]^T
\end{align}
where $\forall j$, we have
\begin{align}
    \frac{\partial L_\rho (\alpha,\lambda)}{\partial \lambda_j}=h_j(\alpha).
\end{align}

\section{Proof of Theorem \ref{thm:kkt_limit}}
\label{app:proof_kkt_limit}

\renewcommand{\thetheorem}{\ref{thm:kkt_limit}}

\begin{theorem}[KKT Property of Limit Points]
Under Assumption \ref{assump:primal_conv}, if the sequence $\{(\alpha^t,\lambda^t)\}$ generated by Algorithm \ref{algo:ncb} converges to a certain point $(\hat{\alpha}^*,\hat{\lambda}^*)$, then this limit point admits a KKT point of the NCB problem in Eq. \eqref{equ:ncb_alpha_simplify}, i.e.,
\begin{align}
 &(\text{stationary})\qquad\sum_{i\in[N]}\bigg[   \nabla_\alpha R_i(\hat{\alpha}^*)+\hat{\lambda}^*_i\nabla_\alpha h_i(\hat{\alpha}^*)\bigg]=0\notag\\
& (\text{feasibility})\qquad   h_i(\hat{\alpha}^*)=0,\forall i.\notag
\end{align}
\end{theorem}

\begin{proof}
As the sequence $\{(\alpha^t,\lambda^t)\}$ converges, from the update rule of $\lambda^t$, we can derive:
\begin{align}
    &\lambda^{t+1}_i=\lambda^{t}_i-\rho h_i(\alpha)\notag\\
    \Rightarrow \qquad&\lim_{t\rightarrow\infty}h_i(\alpha)=\lim_{t\rightarrow \infty}\frac{{\lambda}^t-{\lambda}^{t+1}}{\rho}=\frac{\hat{\lambda}^*-\hat{\lambda}^*}{\rho}=0.
\end{align}
This indicates that the limit point is a feasible point of the NCB problem in Eq. \eqref{equ:ncb_alpha_simplify}, i.e.,
\begin{align}
\label{equ:fea}
    (\text{feasibility})\qquad h_i(\hat{\alpha}^*)=0,\quad\forall i.
\end{align}
Moreover, under Assumption \ref{assump:primal_conv}, it holds that:
\begin{align}
    \lim_{t\rightarrow\infty}\nabla_\alpha L_\rho(\alpha^{t+1},\lambda^t)&=\lim_{t\rightarrow\infty}\sum_{i\in [N]}\bigg[\nabla_\alpha R_i(\alpha^{t+1})+ \bigg(\lambda_i^t-\rho \underbrace{h_i(\alpha)}_{=0}\bigg)
    \nabla_\alpha h_i(\alpha^{t+1})\bigg]\notag\\
    &=\sum_{i\in [N]}\bigg[\nabla_\alpha R_i(\hat{\alpha}^{*})+ \hat{\lambda}_i^*
    \nabla_\alpha h_i(\hat{\alpha}^{*})\bigg]\notag\\
    &=0.
\end{align}
This indicates that  $(\hat{\alpha}^*,\hat{\lambda}^{*})$ pair satisfies the stationary condition, i.e.,
\begin{align}
\label{equ:sationary}
  (\text{stationary})\quad\quad  \underbrace{\sum_{i\in [N]}\bigg[\nabla_\alpha R_i(\hat{\alpha}^*)+ \hat{\lambda}^{*}_i
    \nabla_\alpha h_i(\hat{\alpha}^*)\bigg]}_{\text{the gradient of the Lagrangian function of Eq. \eqref{equ:ncb_alpha_simplify}}}=0.
\end{align}
Eq. \eqref{equ:sationary} and Eq. \eqref{equ:fea} complete the proof.
\end{proof}

\section{Proof of Theorem \ref{thm:convergence_guarantee}}
\label{app:proof_conv}

Before diving into the proof of Theorem \ref{thm:convergence_guarantee}, we provide three lemmas that will be used later.

\renewcommand{\thetheorem}{G.1}
\begin{lemma}[Invertible Jacobian Matrix]
\label{lemma:invertible_J}
    Under Assumption \ref{assump:self_dominance}, the Jacobian matrix $J_H(\alpha)$ is invertible. 
\end{lemma}
\begin{proof}
    Under Assumption \ref{assump:self_dominance}, we can show that $J_H(\alpha)$ is a strictly diagonal dominant matrix from Eq.~\eqref{equ:nabla_h}, i.e., $\forall i$,
\begin{align}
    \bigg|\frac{\partial h_i(\alpha)}{\partial \alpha_i}\bigg|&=\bigg|-\underbrace{\bigg(1-\frac{B_i-C_i(\alpha)}{Z_i}\bigg)}_{\triangleq a_i}\frac{\partial C_i(\alpha)}{\partial\alpha_j}-\underbrace{\bigg(1-\frac{A-\alpha_i}{Z_i}\bigg)}_{\triangleq b_i}\delta_{ij}\bigg|\notag\\
    &=|a_i|\bigg|\frac{\partial C_i(\alpha)}{\partial\alpha_i}\bigg|+|b_i|\notag\\
    &>|a_i|\sum_{j\neq i}\bigg|\frac{\partial C_i(\alpha)}{\partial \alpha_j}\bigg|\notag\\
    &=\sum_{j\neq i} \bigg|\frac{\partial h_i(\alpha)}{\partial\alpha_j}\bigg|.
\end{align}
With the Lévy–Desplanques theorem \cite{horn2012matrix}, a strictly diagonal dominant matrix $J_H(\alpha)$ is invertible. 
\end{proof}

\renewcommand{\thetheorem}{G.2}
\begin{lemma}[Invertible Hessian Matrix]
\label{lemma:invertible_H}
If $\sigma_\text{min}^2(J_H(\alpha))>\bar{H}_1$ and the penalty factor $\rho$ is large enough such that 
\begin{align}
    \rho > \frac{\Lambda_{H_0}}{\sigma_\text{min}^2(J_H(\alpha))-\bar{H}_1},
\end{align}
where 
\begin{itemize}
    \item $H_0(\alpha,\lambda)\triangleq\sum_{i\in[N]}[\nabla_{\alpha\alpha}^2R_i(\alpha,\lambda)+\lambda_i\nabla_{\alpha\alpha}^2h_i(\alpha)]$ 
    , and $\Lambda_{H_0}$ denotes the maximum eigenvalue of $H_0$;
    \item $H_1(\alpha,\lambda)\triangleq \sum_{i\in [N]}h_i(\alpha)\nabla^2_{\alpha\alpha}h_i(\alpha)$, and $\bar{H}_1$ denotes the upper bound of its norm.
\end{itemize}
then, under Assumption \ref{assump:self_dominance}, the Hessian matrix $\nabla^2_{\alpha\alpha}L_\rho(\alpha,\lambda)$ negative definite and thereby, is invertible. 
    
\end{lemma}
\begin{proof}
    The Hessian matrix of the augmented Lagrangian function $L_\rho(\alpha,\lambda)$ with respect to $\alpha$ can be expressed as:
    \begin{align}
        \nabla^2_{\alpha\alpha}L_\rho(\alpha,\lambda)=\underbrace{\sum_{i\in[N]}\bigg[\nabla_{\alpha\alpha}^2R_i(\alpha,\lambda)+\lambda_i\nabla_{\alpha\alpha}^2h_i(\alpha)\bigg]}_{\triangleq H_0(\alpha,\lambda)}-\rho \bigg[J_H(\alpha)^TJ_H(\alpha)+\underbrace{\sum_{i\in [N]}h_i(\alpha)\nabla^2_{\alpha\alpha}h_i(\alpha)}_{\triangleq H_1(\alpha,\lambda)}\bigg],
    \end{align}
To verify that the Hessian matrix $\nabla_{\alpha\alpha}^2L_\rho(\alpha,\lambda)$ is negative definite, we show that for any non-zero vector $x$:
\begin{align}
    x^T(\nabla_{\alpha\alpha}^2L_\rho(\alpha,\lambda))x&=x^TH_0(\alpha,\lambda)x-\rho \bigg[\|J_H(\alpha)x\|_2^2 +x^TH_1(\alpha,\lambda)x\bigg]\notag\\
    &\le \bigg[\Lambda_{H_0}-\rho\bigg(\sigma_\text{min}^2(J_H)-\underbrace{\sum_{i\in [N]}|h_i(\alpha)|\|\nabla^2_{\alpha\alpha}h_i(\alpha)\|_2}_{\triangleq \bar{H}_1}\bigg)\bigg]\|x\|^2_2,
\end{align}
where $\Lambda_{H_0}$ is the largest eigenvalue of $H_0(\alpha,\lambda)$, $\sigma_\text{min}^2(J)$ is the smallest singular value of $J_H(\alpha)$ which is positive due to Lemma \ref{lemma:invertible_J}, and $\bar{H}_1$ is the upper bound of the norm of $H_1(\alpha,\lambda)$.
Here, as $J_H(\alpha)$ is invertible, we have $\sigma_\text{min}(J_H(\alpha))>0$.
Therefore, when $\sigma_\text{min}(J_H(\alpha))^2>\bar{H}_1$, the Hessian matrix is negative definite, i.e., $x^T(\nabla_{\alpha\alpha}^2L_\rho(\alpha,\lambda))x<0$, if:
\begin{align}
    \rho > \frac{\Lambda_{H_0}}{\sigma_\text{min}(J_H(\alpha))^2-\bar{H}_1}.
\end{align}
This indicates that $J_H(\alpha)$ is negative definite, which completes the proof.
\end{proof}

In addition, we provide a lemma on the local dual function $d(\lambda)$ that will be used later.
Specifically, the local dual function $d(\lambda)$ is defined as:
\begin{align}
    d(\lambda)=L_\rho(\alpha^*_l(\lambda), \lambda),
\end{align}
where $\alpha^*_l(\lambda)$ is the local optima of $L_\rho(\alpha, \lambda)$ that satisfies the first-order optimality condition, i.e., $\nabla_\alpha L_\rho(\alpha^*_l(\lambda), \lambda)=0$.
The gradient of $d(\lambda)$ can be derived as:
\begin{align}
    \nabla_\lambda d(\lambda)=\frac{\partial L_\rho(\alpha^*_l(\lambda), \lambda)}{\partial \lambda}+\underbrace{(\nabla_\alpha L_\rho(\alpha^*_l(\lambda), \lambda))^T}_{=0}\frac{\partial\alpha^*_l(\lambda)}{\partial\lambda}=\frac{\partial L_\rho(\alpha^*_l(\lambda), \lambda)}{\partial \lambda}.
\end{align}
This is commonly known as the Envelope Theorem \cite{milgrom2002envelope}.
Specifically, we have:
\begin{align}
    \frac{\partial d(\lambda)}{\partial\lambda_i}=h_i(\alpha^*_l(\lambda)).
\end{align}
We show that $\nabla_\lambda d(\lambda)$ is a Lipschitz continuous function.
\renewcommand{\thetheorem}{G.3}
\begin{lemma}[Lipschitz Continuity of $\nabla_\lambda d(\lambda)$]
\label{lemma:lip_d}
Under Assumption \ref{assump:self_dominance}, 
if $\sigma_\text{min}^2(J_H(\alpha))>\bar{H}_1$ and the penalty factor $\rho$ is large enough such that $\rho > \frac{\Lambda_{H_0}}{\sigma_\text{min}^2(J_H(\alpha))-\bar{H}_1}$,
then the local dual function $\nabla_\lambda d(\lambda)$ is an $L$-Lipschitz continuous function, i.e., 
\begin{align}
    \|\nabla_\lambda d(\lambda_1)-\nabla_\lambda d(\lambda_2)\|_2\le L \|\lambda_1-\lambda_2\|_2,
\end{align}
holds $\forall \lambda_1, \lambda_2$, 
and
\begin{align}
    L\triangleq 2N\bigg(1+KV+\frac{ AKV^2}{\tau}\bigg)\frac{L_c}{\mu}, 
\end{align}
where $L_c$ and $1/\mu$ are the upper bounds of $\|\nabla_{\alpha\lambda}^2L_\rho\|_2$ and $\|(\nabla_{\alpha\alpha}^2 L_\rho)^{-1}\|_2$, respectively. 
\end{lemma}
\begin{proof}
We examine the upper bound $\|\nabla_\lambda d(\lambda_1)-\nabla_\lambda d(\lambda_2)\|_2, \forall \lambda_1, \lambda_2$. Specifically, 
\begin{align}
    \bigg|\frac{\partial d(\lambda_1)}{\partial\lambda_i}-\frac{\partial d(\lambda_2)}{\partial\lambda_i}\bigg|&=|h_i(\alpha^*_l(\lambda_1))-h_i(\alpha^*_l(\lambda_2))|\notag\\
    &\le \bigg|C_i(\alpha^*_l(\lambda_1))-C_i(\alpha^*_{l}(\lambda_2))\bigg| + \bigg|\alpha^*_{l,i}(\lambda_1)-\alpha^*_{l,i}(\lambda_2)\bigg| + \bigg|Z_i(\alpha^*_l(\lambda_1))-Z_i(\alpha^*_l(\lambda_2))\bigg|
\end{align}
We next examine the Lipschitz property of $C_i$ and $V_i$ terms on the right-hand side, respectively. 

\textbf{(1) Lipschitz of Cost $C_i$.}
\begin{align}
\label{equ:l_1_l_2}
   \bigg|C_i(\alpha^*_l(\lambda_1))-C_i(\alpha^*_l(\lambda_2))\bigg|&\le \sum_{k\in [K]}\bigg|p_{i,k}(\alpha^*_l(\lambda_1))m_{i,k}(\alpha^*_l(\lambda_1))-p_{i,k}(\alpha^*_l(\lambda_2))m_{i,k}(\alpha^*_l(\lambda_2))\bigg|\notag
    \\
   & \le\sum_{k\in [K]} \underbrace{\bigg|p_{i,k}(\alpha^*_l(\lambda_1))\bigg(m_{i,k}(\alpha^*_l(\lambda_1))-m_{i,k}(\alpha_l^*(\lambda_2))\bigg)\bigg|}_{\triangleq l_1}+\notag\\
   &\qquad\;\; \quad\underbrace{\bigg|\bigg(p_{i,k}(\alpha^*_l(\lambda_1))-p_{i,k}(\alpha^*_l(\lambda_2))\bigg)m_{i,k}(\alpha_l^*(\lambda_2))\bigg|}_{\triangleq l_2}.
\end{align}
We next examine $l_1$ and $l_2$, respectively. Specifically, we have: 
\begin{align}
    l_1\le p_{i,k}(\alpha^*_l(\lambda_1))\bigg|m_{i,k}(\alpha^*_l(\lambda_1))-m_{i,k}(\alpha_l^*(\lambda_2))\bigg|\le \bigg|m_{i,k}(\alpha^*_l(\lambda_1))-m_{i,k}(\alpha_l^*(\lambda_2))\bigg|.
\end{align}
We check the Lipschitz constant through the gradient of the market price $m_{i,k}$. Specifically, from Eq.~\eqref{equ:partial_m}, we have:
\begin{align}
    \|\nabla_\alpha m_{i,k}\|_1&=\sum_{j\in[N]} \bigg|p^{-i}_{j,k}v_{j,k}\bigg[1+\frac{\alpha_j v_{j,k}-m_{i,k}}{\tau}\bigg](1-\delta_{ij})\bigg|\notag\\
    &\le \sum_{j\neq i}\bigg|v_{j,k}p^{-i}_{j,k}\bigg|+\sum_{j\neq i}\bigg|v_{j,k}p^{-i}_{j,k}\frac{\alpha_j v_{j,k}-m_{i,k}}{\tau}\bigg|\notag\\
    &\le V\sum_{j\neq i}p^{-i}_{j,k}+V\underbrace{\sum_{j\neq i} p^{-i}_{j,k}\bigg|\frac{\alpha_j v_{j,k}-m_{i,k}}{\tau}\bigg|}_{\text{Weighted Mean Absolute Deviation of $X$}}\notag\\
    &\le V+V\sqrt{\text{Var}(X)}\notag\\
    &\le V + V\frac{AV}{2\tau}\notag\\
    &=V\bigg(1+\frac{AV}{2\tau}\bigg),
\end{align}
where we let $X$ be a random variable with $N-1$ possible values $\{\frac{\alpha_1v_{1,k}}{\tau},\cdots,\frac{\alpha_{i-1}v_{i-1,k}}{\tau},\frac{\alpha_{i+1}v_{i+1,k}}{\tau},\cdots,\frac{\alpha_Nv_{N,k}}{\tau}\}$, and $\{p^{-i}_{j,k}\}$ denotes the corresponding probability.
Therefore, we have, $\forall \alpha_1,\alpha_2$:
\begin{align}
|m_{i,k}(\alpha_1)-m_{i,k}(\alpha_2)|\le \|\nabla_\alpha m_{i,k}(\alpha)\|_1\|\alpha_1-\alpha_2\|_\infty\le V\bigg(1+\frac{AV}{2\tau}\bigg)\|\alpha_1-\alpha_2\|_2.
\end{align}
This indicates that:
\begin{align}
    l_1\le V\bigg(1+\frac{AV}{2\tau}\bigg)\|\alpha_l^*(\lambda_1)-\alpha_l^*(\lambda_2)\|_2.
\end{align}

In addition, we have:
\begin{align}
\label{equ:l_2_init}
    l_2\le \max_{j\neq i}\alpha^*_{2,j}\max_{j\neq i}v_{j,k}\bigg|p_{i,k}(\alpha^*_l(\lambda_1))-p_{i,k}(\alpha^*_l(\lambda_2))\bigg|\le AV\bigg|p_{i,k}(\alpha^*_l(\lambda_1))-p_{i,k}(\alpha^*_l(\lambda_2))\bigg|.
\end{align}
For the winning probability $p_{i,k}$, we check its Lipschitz constant through its gradient. Specifically, from Eq.~\eqref{equ:nabla_p}, we have
\begin{align}
\label{equ:p_l}
    \|\nabla_\alpha p_{i,k}(\alpha)\|_1=&\sum_{j\in [N]} \bigg|\frac{p_{i,k}v_{j,k}}{\tau}(\delta_{ij}-p_{j,k})\bigg|\notag\\
    &=p_{i,k}\bigg[\bigg|\frac{v_{j,k}}{\tau}(1-p_{j,k})\bigg|+\sum_{j\neq i}\frac{1}{\tau}|v_{j,k}p_{j,k}|\bigg]\notag\\
    &\le p_{i,k} \bigg[\frac{V}{\tau}(1-p_{i,k})+\frac{V}{\tau}(1-p_{i,k})\bigg]\notag\\
    &=\frac{2V}{\tau}p_{i,k}(1-p_{i,k})\notag\\
    &\le \frac{V}{2\tau}
\end{align}
Therefore, combining Eq. \eqref{equ:l_2_init} and Eq. \eqref{equ:p_l}, we have:
\begin{align}
    l_2\le \frac{ AV^2}{2\tau}\|\alpha^*_{l}(\lambda_1)-\alpha^*_{l}(\lambda_2)\|_2.
\end{align}
Substituting $l_1$ and $l_2$ in Eq. \eqref{equ:l_1_l_2}, we have:
\begin{align}
\label{equ:lip_c}
   \bigg|C_i(\alpha^*_l(\lambda_1))-C_i(\alpha^*_l(\lambda_2))\bigg|\le KV\bigg(1+\frac{ AV}{\tau}\bigg)\|\alpha^*_{l}(\lambda_1)-\alpha^*_{l}(\lambda_2)\|_2
\end{align}

\textbf{(2) Lipschitz Property of $Z_i$.} Let $r_i\triangleq \sqrt{x_i^2+y_i^2}$, where $x_i\triangleq B_i-C_i(\alpha)$ and $y_i\triangleq A-\alpha_i$. We have:
\begin{align}
\label{equ:lip_h}
    |Z_i(\alpha_1)-Z_i(\alpha_2)|&=|\sqrt{r_i^2(\alpha_1)+\epsilon}-\sqrt{r_i^2(\alpha_2)+\epsilon}|\notag\\
    &=\frac{|r_i^2(\alpha_1)-r_i^2(\alpha_2)|}{\sqrt{r_i^2(\alpha_1)+\epsilon}+\sqrt{r_i^2(\alpha_2)+\epsilon}}\notag\\
    &=|r_i(\alpha_1)-r_i(\alpha_2)|\underbrace{\frac{|r_i(\alpha_1)+r_i(\alpha_2)|}{\sqrt{r_i^2(\alpha_1)+\epsilon}+\sqrt{r_i^2(\alpha_2)+\epsilon}}}_{\le 1}\notag\\
    &\le |r_i(\alpha_1)-r_i(\alpha_2)|\notag\\
    &=\underbrace{\bigg|\sqrt{x_i^2(\alpha_1)+y_i^2(\alpha_1)}-\sqrt{x_i^2(\alpha_2)+y_i^2(\alpha_2)}\bigg|}_{\text{The difference between two sides of a triangle is smaller than the third}}\notag\\
    &\le \sqrt{(x_i(\alpha_1)-x_i(\alpha_2))^2+(y_i(\alpha_1)-y_i(\alpha_2))^2}\notag\\
    &=\sqrt{(C_i(\alpha_1)-C_i(\alpha_2))^2+(\alpha_{1,i}-\alpha_{2,i})^2}\notag\\
    &\le \bigg|C_i(\alpha_1)-C_i(\alpha_2)\bigg| + \bigg|\alpha_{1,i}-\alpha_{2,i}\bigg|\notag\\
    &\le \bigg(KV(1+\frac{AV}{\tau})+1\bigg)\|\alpha_1-\alpha_2\|_2,
\end{align}
where we leverage the Lipschitz property of the cost $C_i$ in Eq.~\eqref{equ:lip_c}.

\textbf{(3) Lipschitz property of $h_i$.} Combining Eq.~\eqref{equ:lip_c} and \eqref{equ:lip_h}, we have:
\begin{align}
      \bigg|\frac{\partial d(\lambda_1)}{\partial\lambda_i}-\frac{\partial d(\lambda_2)}{\partial\lambda_i}\bigg|\le 2\bigg(1+KV+\frac{AKV^2}{\tau}\bigg)\|\alpha_l^*(\lambda_1)-\alpha_l^*(\lambda_1)\|_2.
\end{align}

\textbf{(4) Relationship between $\alpha^*_l(\lambda)$ and $\lambda$.}
We next check the relationship between $\|\alpha_l^*(\lambda_1)-\alpha_l^*(\lambda_1)\|_2$ and $\|\lambda_1-\lambda_2\|_2$.
Specifically, as $\alpha^*_l(\lambda)$ is the local optima of $L_\rho(\alpha,\lambda)$ given $\lambda$, it holds that:
\begin{align}
    &\nabla_\alpha L_\rho(\alpha^*_l(\lambda), \lambda)=0\notag\\
   (\text{take the derivative of $\lambda$})\;\; {\Rightarrow}
    \quad\quad&\nabla_{\alpha\lambda}^2L_\rho(\alpha^*_l(\lambda), \lambda) + \nabla_{\alpha\alpha}^2 L_\rho(\alpha^*_l(\lambda), \lambda)\frac{d\alpha^*_l(\lambda)}{d \lambda}=0\notag\\
    \Rightarrow\qquad&\frac{d \alpha^*_l(\lambda)}{d \lambda}=-\bigg(\nabla_{\alpha\alpha}^2 L_\rho\bigg)^{-1}\nabla_{\alpha\lambda}^2L_\rho.
    \label{equ:implicit_function}
\end{align}
where the invertible of $\nabla^2_{\alpha\alpha}L_\rho$ is guaranteed by  Lemma \ref{lemma:invertible_H}.
Based on Eq. \eqref{equ:implicit_function}, we have $\forall \lambda_1,\lambda_2$:
\begin{align}
\label{equ:alpha_lambda}
    \|\alpha^*_l(\lambda_1)-\alpha_l^*(\lambda_2)\|_2&\le\bigg(\max_{\alpha,\lambda}\|(\nabla_{\alpha\alpha}^2 L_\rho)^{-1}\|_2\|\nabla_{\alpha\lambda}^2L_\rho\|_2\bigg) \|\lambda_1-\lambda_2\|_2\notag\\
    &\le \frac{L_c}{\mu}\|\lambda_1-\lambda_2\|_2,
\end{align}
where $L_c$ and $1/\mu$ are the upper bounds of $\|\nabla_{\alpha\lambda}^2L_\rho\|_2$ and $\|(\nabla_{\alpha\alpha}^2 L_\rho)^{-1}\|_2$, respectively. 
Combining Eq. \eqref{equ:lip_h} and Eq. \eqref{equ:alpha_lambda}, we have $\forall \lambda_1,\lambda_2$:
\begin{align}
    \bigg|\frac{\partial d(\lambda_1)}{\partial\lambda_i}-\frac{\partial d(\lambda_2)}{\partial\lambda_i}\bigg|\le 2\bigg(1+KV+\frac{ AKV^2}{\tau}\bigg)\frac{L_c}{\mu} \|\lambda_1-\lambda_2\|_2.
\end{align}
Then, it holds that:
\begin{align}
     \|\nabla_\lambda d(\lambda_1)-\nabla_\lambda d(\lambda_2)\|_2&=\sqrt{\sum_{i\in [N]}\bigg(\frac{\partial d(\lambda_1)}{\partial\lambda_i}-\frac{\partial d(\lambda_2)}{\partial\lambda_i}\bigg)^2}\notag\\
     &\le \sum_{i\in [N]} \bigg|\frac{\partial d(\lambda_1)}{\partial\lambda_i}-\frac{\partial d(\lambda_2)}{\partial\lambda_i}\bigg|\notag\\
     &\le \underbrace{2N\bigg(1+KV+\frac{ AKV^2}{\tau}\bigg)\frac{L_c}{\mu} }_{\triangleq L}\|\lambda_1-\lambda_2\|_2,
\end{align}
which indicates that $\nabla_\lambda d(\lambda)$ is an $L$-Lipschitz function with respect to $\lambda$.
This completes the proof.
\end{proof}

Based on Lemmas \ref{lemma:invertible_J}, \ref{lemma:invertible_H} and \ref{lemma:lip_d}, we here give the proof of Theorem \ref{thm:convergence_guarantee} as follows.
\renewcommand{\thetheorem}{\ref{thm:convergence_guarantee}} 
\begin{theorem}[Global Convergence]
Under Assumptions \ref{assump:primal_conv} and \ref{assump:self_dominance}, if $\sigma^2_\text{min}(J_H(\alpha))>\bar{H}_1$, then with sufficiently large penalty factor that satisfies
\begin{align}
    \rho > \frac{\Lambda_{H_0}}{\sigma^2_\text{min}(J_H(\alpha))-\bar{H}_1},
\end{align}
the sequence $\{(\alpha^t,\lambda^t)\}$ generated by Algorithm \ref{algo:ncb} is guaranteed to converge as $t \rightarrow \infty$.
Note that here $\sigma_\text{min}(J_H(\alpha))$ denotes the minimal singular value of $J_H(\alpha)$, and $\Lambda_{H_0}$ is the maximum eigenvalue of the Hessian matrix of the Lagrangian function of Eq. \eqref{equ:ncb_alpha_simplify}, and  
$\bar{H}_1$ denotes the upper bound of $\|\sum_{i\in [N]}h_i(\alpha)\nabla^2_{\alpha\alpha}h_i(\alpha)\|_2$.
\end{theorem}

\begin{proof}
Recall that the dual update is:
\begin{align}
\label{equ:lambda_update}
    \lambda^{t+1}\leftarrow\lambda^t -\eta\underbrace{h_i(\alpha^{t+1}(\lambda^t))}_{=\frac{\partial d(\lambda^t)}{\partial \lambda_i}},
\end{align}
which is exactly conducting the gradient descent on the local dual function $d(\lambda)$.
Recall that from Lemma \ref{lemma:lip_d}, if $\sigma^2_\text{min}(J_H(\alpha))>\bar{H}_1$ and the penalty factor is sufficiently large, i.e., 
\begin{align}
    \rho > \frac{\Lambda_{H_0}}{\sigma^2_\text{min}(J_H(\alpha))-\bar{H}_1},
\end{align}
then the gradient of the local dual function $\nabla_\lambda d(\lambda)$ is $L$-Lipschitz.
With the Descent Lemma \cite{bertsekas1997nonlinear}, it holds that:
\begin{align}
\label{equ:d_decrease}
    d(\lambda^{t+1})\le d(\lambda^t)-\eta\bigg(1-\frac{L\eta}{2}\bigg)\|\nabla_\lambda d(\lambda^t)\|^2_2.
\end{align}
This indicates that if $\eta < \frac{2}{L}$, then $d(\lambda^{t+1})\le d(\lambda^t)$. Note that $d(\lambda)$ is lower bounded at any finite $\lambda$ since $R(\alpha)$ and $C(\alpha
)$ are both bounded with $\alpha\in [0,A]^N$.
This indicates that the sequence $\{d(\lambda^t)\}$ is a bounded monotonically decrease sequence, which is guaranteed to converge.
From Eq.~\eqref{equ:d_decrease}, we know that if $d(\lambda)$ converges, then: 
\begin{align}
\nabla_\lambda d(\lambda^t)=0.    
\end{align}
From Eq.~\eqref{equ:lambda_update}, we know that the update step of $\lambda^t$ is zero when $\nabla_\lambda d(\lambda^t)=0$, indicating a convergence of $\{\lambda^t\}$ sequence. 
From Eq.~\eqref{equ:alpha_lambda}, we know that:
\begin{align}
     \|\alpha^*_l(\lambda^{t+1})-\alpha_l^*(\lambda^t)\|_2\le \frac{L_c}{\mu}\|\lambda^{t+1}-\lambda^t\|_2\le 0,
\end{align}
where $\{\alpha^t\}$ sequence in Algorithm \ref{algo:ncb} is exactly $\{\alpha^*_l(\lambda^t)\}$ sequence due to Assumption \ref{assump:primal_conv}.
From the above equation, we have $\lim_{t\rightarrow\infty} \|\alpha^{t+1}-\alpha^t\|_2=0$, which indicates that $\{\alpha^t\}$ sequence converges.
Therefore, the sequence $\{(\alpha^t,\lambda^t)\}$ generated by Algorithm \ref{algo:ncb} converges.

\section{Derivations in Complexity Analysis}
 \label{app:complex_derivation}
We here examine the constraint gradient term:
    \begin{align}
       \sum_{i\in[N]}E_i  \frac{\partial h_i(\alpha)}{\partial \alpha_j}&=\sum_{i\in [N]} E_i\underbrace{\bigg(-1+\frac{B_i-C_i}{Z_i}\bigg)}_{\triangleq a_i}\frac{\partial C_i}{\partial\alpha_j}
   +E_i\underbrace{\bigg(-1+\frac{A-\alpha_i}{Z_i}\bigg)}_{\triangleq b_i}\delta_{ij}\notag\\
   &=\sum_{i\in [N]}a_iE_i\sum_{k\in [K]}\bigg[p_{i,k}\underbrace{p^{-i}_{j,k}}_{=\frac{p_{j,k}}{1-p_{i,k}}}v_{j,k}\bigg[1+\frac{\alpha_j v_{j,k}-m_{i,k}}{\tau}\bigg](1-\delta_{ij})+m_{i,k}\frac{p_{i,k}v_{j,k}}{\tau}(\delta_{ij}-p_{j,k})\bigg]+b_jE_j\notag\\
   &=\sum_{k\in [K]}p_{j,k}v_{j,k}\underbrace{\sum_{i\neq j}a_iE_ip_{i,k}\frac{1}{1-p_{i,k}}\bigg[1+\frac{\alpha_j v_{j,k}-m_{i,k}}{\tau}\bigg]}_{\triangleq I_1}\notag\\
   &\quad+\sum_{k\in [K]}\bigg[a_jE_jm_{j,k}\frac{p_{j,k}v_{j,k}}{\tau}-p_{j,k}v_{j,k}\underbrace{\sum_{i\in [N]}a_iE_im_{i,k}\frac{p_{i,k}}{\tau}}_{\triangleq U_k}\bigg]+b_jE_j.
    \end{align}
We examine $I_1$ in the following:
\begin{align}
    I_1&=\bigg(1+\frac{\alpha_jv_{j,k}}{\tau}\bigg)\underbrace{\sum_{i\neq j}a_iE_ip_{i,k}\frac{1}{1-p_{i,k}}}_{=Q_k-a_jE_j\frac{p_{j,k}}{1-p_{j,k}}}-\frac{1}{\tau}\underbrace{\sum_{i\neq j}a_iE_ip_{i,k}\frac{1}{1-p_{i,k}}m_{i,k}}_{=Q_k'-a_jE_j\frac{p_{j,k}}{1-p_{j,k}}m_{j,k}}\notag\\
    &=\bigg(1+\frac{\alpha_jv_{j,k}}{\tau}\bigg)(Q_k-a_jE_j\frac{p_{j,k}}{1-p_{j,k}})-\frac{1}{\tau}(Q_k'-a_jE_j\frac{p_{j,k}}{1-p_{j,k}}m_{j,k}),
\end{align}
where 
\begin{align}
    Q_k\triangleq\sum_{i\in [N]} a_iE_i\frac{p_{i,k}}{1-p_{i,k}}, \quad Q'_k\triangleq\sum_{i\in [N]} a_iE_i\frac{p_{i,k}}{1-p_{i,k}}m_{i,k}
\end{align}
\end{proof}

\section{Experiment Baselines}
\label{app:baselines}
\begin{itemize}
	\item USCB \cite{USCB} is a state-of-the-art SAB algorithm. It leverages the DDPG algorithm to learn the auto-bidding policies.
	\item V-CQL \cite{SORL} is a state-of-the-art SAB algorithm leveraging offline RL methods.  
	\item MAAB \cite{MAAB} is a state-of-the-art MAB algorithm. It is designed to balance the trade-off between cooperation and competition among advertisers using RL methods, aiming to achieve high social welfare and revenue, which inherently aligns with the NCB problem. Specifically, MAAB designs a credit assignment method to incentivize social welfare among agents and develops a bar agent to ensure the platform's revenue. However, these are essentially heuristic methods that lack theoretical foundations to guarantee a balanced trade-off. 
	\item DCMAB \cite{DCMAB} is a popular MAB algorithm that aims to balance the trade-off between cooperation and competition between advertisers based on RL methods, inherently aligning with the NCB problem. However, DCMAB directly adopts MADDPG \cite{MADDPG}, which also lacks theoretical foundations to ensure a balance between cooperation and competition. 
\end{itemize}


\end{document}